\documentclass{article}

\usepackage[hyphens]{url}

\usepackage{microtype}
\usepackage{graphicx}
\graphicspath{ {./} }
\usepackage{subfigure}
\usepackage{booktabs} % for professional tables
\usepackage{mdwlist} % squeezing tables

\usepackage[breaklinks]{hyperref}

\usepackage{eqparbox}
 \usepackage[accepted]{myicml}

\usepackage{amsmath,bm,graphicx,psfrag,epsf,enumerate,placeins,float,sectsty,amstext,relsize,wrapfig,listings,subfigure,amsfonts,amsthm,mathabx,enumitem,hyperref,grffile,colortbl,dsfont,bbm,setspace, float, xcolor}

\usepackage{balance}
\usepackage{mdframed} %nice frames for code snippet
\usepackage[percent]{overpic}
\usepackage{adjustbox}

\usepackage{multibib} % allows 2 bibliographies
\newcites{si}{Additional References for the Appendix}

\newcommand{\beginsupplement}{ % use to mark beginning of supplementary section. 
        \setcounter{section}{0}
        \renewcommand{\thesection}{S\arabic{section}} %
         \renewcommand{\thesubsection}{\thesection.\arabic{subsection}}
        \setcounter{table}{0}
        \renewcommand{\thetable}{S\arabic{table}} %
        \setcounter{figure}{0}
        \renewcommand{\thefigure}{S\arabic{figure}} %
     }

\definecolor{LightGray}{gray}{0.9}

\definecolor{lighter-gray}{gray}{0.95} %the shade of grey that stack exchange uses

\usepackage{xcolor}
 
\definecolor{codegreen}{rgb}{0,0.6,0}
\definecolor{codegray}{rgb}{0.5,0.5,0.5}
\definecolor{codepurple}{rgb}{0.58,0,0.82}
\definecolor{backcolour}{rgb}{0.95,0.95,0.92}
 
\lstdefinestyle{mystyle}{
    basicstyle=\fontsize{8}{8}\selectfont\ttfamily,
    commentstyle=\color{codegreen},
    keywordstyle=\color{magenta},
    numberstyle=\tiny\color{codegray},
    stringstyle=\color{codepurple},
    breakatwhitespace=false,         
    breaklines=true,                 
    captionpos=b,                    
    keepspaces=true,                 
    numbers=left,                    
    numbersep=5pt,                  
    showspaces=false,                
    showstringspaces=false,
    showtabs=false,                  
    tabsize=2
}
 
\allowdisplaybreaks  % allow align to cross pages

\usepackage{tikz}
\usetikzlibrary{calc, arrows, arrows.meta,positioning}

\newcounter{daggerfootnote}

\newcommand{\papertitle}{AutoGluon-Tabular: Robust and
  Accurate AutoML for Structured Data}

\icmltitlerunning{\papertitle}

\newcommand{\gitlink}{\href{https://github.com/awslabs/autogluon}{\nolinkurl{github.com/awslabs/autogluon}}}
\newcommand{\website}{\href{https://autogluon.mxnet.io}{\nolinkurl{autogluon.mxnet.io}}}
\newcommand{\benchmarkgitlink}{\href{https://github.com/Innixma/autogluon-benchmarking}{\nolinkurl{github.com/Innixma/autogluon-benchmarking}}}

\begin{document}

\twocolumn[
\icmltitle{\papertitle}



\icmlsetsymbol{equal}{*}

\begin{icmlauthorlist}
\icmlauthor{Nick Erickson}{equal,am} 
\icmlauthor{Jonas Mueller}{equal,am}
\icmlauthor{Alexander Shirkov}{am}
\icmlauthor{Hang Zhang}{am}
\icmlauthor{Pedro Larroy}{am} \\
% \icmlauthor{Chenguang Wang}{am}
\icmlauthor{Mu Li}{am} 
\icmlauthor{Alexander Smola}{am} 
\end{icmlauthorlist}   
   
\icmlaffiliation{am}{Amazon}
% \icmlaffiliation{am}{Amazon Web Services}
% \icmlaffiliation{alexa}{Amazon Alexa}

 \icmlcorrespondingauthor{Nick Erickson}{\mbox{neerick@amazon.com}}
 \icmlcorrespondingauthor{Jonas Mueller}{\mbox{jonasmue@amazon.com}}
% \icmlcorrespondingauthor{ \\ NE }{neerick@amazon.com}
%  \icmlcorrespondingauthor{JM }{jonasmue@amazon.com}

\icmlkeywords{AutoML, frameworks, deep learning, model ensembling, stacking}

\vskip 0.3in
]



\printAffiliationsAndNotice{\icmlEqualContribution} % otherwise use the standard text.

 \setcounter{footnote}{1} % ensure open-source footnote doesn't overlap with affiliations.

\begin{abstract}

We introduce AutoGluon-Tabular, % \footnote{\website},
an open-source\footnote{\gitlink} AutoML framework 
%We introduce AutoGluon\footnote{\url{autogluon.mxnet.io}}, an open-source\footnote{\url{github.com/awslabs/autogluon}} AutoML framework 
that requires only a single line of Python to train highly
accurate machine learning models on an % raw, 
unprocessed tabular dataset such as a CSV file.
Unlike existing AutoML frameworks that primarily focus on % model selection and hyperparameter tuning, 
model/hyperparameter selection, 
AutoGluon-Tabular 
succeeds by \emph{ensembling}
multiple models and \emph{stacking} them in multiple layers. 
Experiments reveal that our multi-layer combination of many models offers better use of allocated training time than seeking out the best.

A second contribution is an extensive evaluation of public and
commercial AutoML platforms including TPOT, H2O, AutoWEKA,  auto-sklearn, AutoGluon, % -Tabular,
and Google AutoML Tables. 
% , and SageMaker AutoPilot.  % Not for submission
Tests on a suite of 50 % (supervised)  
classification and regression tasks from Kaggle and the OpenML
AutoML Benchmark reveal that AutoGluon % -Tabular 
is faster, more robust, and much more accurate. We
find that AutoGluon %-Tabular 
often even outperforms the best-in-hindsight combination of all of its competitors.   
In two popular Kaggle competitions, AutoGluon %-Tabular 
beat 99\% of the participating data scientists after merely 4h of training on
the raw data. 

\end{abstract}

\section{Introduction}
\label{sec:intro}

Machine Learning (ML) has advanced significantly over the past decade. This
has led to exciting novel \emph{architectures} for modeling and
techniques for \emph{scaling} estimators to large datasets. It has
also led to a popularization of ML among engineers and
data scientists. However, as state-of-the-art ML techniques
grow in sophistication, it is increasingly difficult even for a
ML expert to incorporate all of the recent best practices into their
modeling. 

AutoML frameworks offer an enticing alternative. For the novice, they remove
many of the barriers of deploying high performance ML models. 
%Equally attractive, 
% For the expert, they offer the potential of implementing best practices such as model selection, stacking, ensembling, hyperparameter tuning, feature engineering to data preprocessing only once and then deploying them many times. 
For the expert, they offer the potential of implementing
best ML practices only once (including strategies for model selection, ensembling, hyperparameter tuning, feature engineering, data preprocessing, data splitting, etc.), and then
being able to repeatedly deploy them. 
This allows experts to scale their knowledge
to many problems without the need for frequent manual intervention.

In this paper, we focus on regression and classification problems with tabular data, drawn IID from some
underlying distribution and stored in a structured table of
values. 
% This setting involves very different design decision than text/image applications. 
%
Numerous AutoML frameworks have
recently emerged to solve this problem,  as evidenced by a flurry of survey
articles \cite{yao2018taking, he2019automl, 
  truong2019towards, guyon2019analysis, gijsbers2019open,
  zoller2019survey}. 
While existing frameworks automate large portions of the supervised
learning pipeline, few of them are able to robustly take raw
data and deliver high-quality predictions without any user input and
without software errors (see Appendix~\ref{sec:errors}). 
Many existing frameworks can only handle numeric data (without missing
values). Such frameworks can thus only be applied to most
datasets after manual preprocessing. Other frameworks can transform raw data to appropriate numeric inputs for ML models, but
require the user to manually specify the type of each variable.

Prior work focused almost exclusively on the task of
Combined Algorithm Selection and Hyperparameter optimization (CASH),
offering strategies to find the best model and its hyperparameters
from a % vast number 
sea of possibilities \cite{thornton2013auto,
  zoller2019survey}.
  % This is much in line with classical risk minimization strategies \cite{vapnik1998statistical}. 
  As this search is % intractable in nonconvex settings,
  typically intractable (nonconvex/nonsmooth), 
   CASH algorithms are
expensive. Their brute-force search % incurs significant computational costs 
expends significant compute evaluating poor model/hyperparameter
configurations that no reasonable data scientist would consider.

In this paper we introduce AutoGluon-Tabular, an easy to use and highly accurate % open-source 
Python library for AutoML with tabular data. In contrast  to prior work focused on CASH, AutoGluon-Tabular performs 
advanced data processing, deep learning, and multi-layer model ensembling. 
It automatically recognizes the data type in each column for robust data preprocessing, including special handling of text fields. AutoGluon fits various models ranging from off-the-shelf boosted trees to customized neural network models. These models are ensembled in a novel way: models are stacked in multiple layers and trained in a layer-wise manner that guarantees raw data can be translated into high-quality predictions within a given time constraint. 
Over-fitting is mitigated throughout this process by splitting the data in various ways with careful tracking of out-of-fold examples.

This paper demonstrates that these admittedly less glamorous aspects of the ML pipeline have a considerable effect on accuracy. Note though, that many of these `tricks of the trade' are well understood in the data science
community, e.g.\ practitioners on Kaggle. To our knowledge, AutoGluon-Tabular is the first framework to codify this extensive set of best practices within a unified framework. 

We additionally introduce several novel extensions that further boost accuracy, including the use of skip connections in both multi-layer stack ensembling and neural network embedding, as well as repeated $k$-fold bagging to curb over-fitting.  
Another contribution of this paper is a thorough experimental study of 6 AutoML frameworks applied to 50 curated datasets that are particularly representative of real ML applications.
The results highlight the substantial impact of these advanced modeling techniques ignored by other AutoML frameworks, which instead allocate their training time budget less resourcefully than AutoGluon.

% It demonstrates the efficiency of best modeling practices other AutoML frameworks ignored, such as modern deep learning and leveraging multi-layer stack ensembling, special handling of text fields, or repeated data splitting with careful tracking of out-of-fold examples.

The rest of this paper is organized as follows.  Section \ref{sec:overview} describes the components of AutoGluon-Tabular. Section \ref{sec:relwork}
surveys the AutoML frameworks we evaluate. Our experimental setup and benchmark results are presented in Section \ref{sec:exp}. The last section provides some concluding remarks.

\section{AutoGluon-Tabular}
\label{sec:overview}

We believe the design of an AutoML framework should adhere to the following principles: 
%\begin{description*}

\textbf{Simplicity.} A user can train a model on the raw data directly without knowing the details about the data and ML models. 

\textbf{Robustness.} The framework can handle a large variety of structured datasets and ensures training succeeds even when some of the individual ML models fail. 

\textbf{Fault Tolerance.} The training can be stopped and resumed at any time. 
Such behavior is preferable when dealing with preemptible (spot) instances on the cloud.

\textbf{Predictable Timing.} It returns the results within the time-budget specified by users.

Next we present each component of AutoGluon-Tabular and discuss how it achieves these principles. 

\subsection{The \texttt{fit} API}

Consider a structured dataset of raw values stored in a CSV file, say \texttt{train.csv}, with the label values to
predict stored in a column named \texttt{class}. Three lines of code are
all that's needed to train and test a model with AutoGluon:
\vskip -0.0em
\begin{lstlisting}[language=Python]
from autogluon import TabularPrediction as task
predictor = task.fit("train.csv", label="class")
predictions = predictor.predict("test.csv")
\end{lstlisting}
\vskip -.1em
Within the call to \texttt{fit()}, AutoGluon automatically: 
preprocesses the raw data, identifies what type of prediction problem
this is (binary, multi-class classification or regression),
% it partitions data to use for model-training vs.\ validation
partitions the data into various folds for model-training vs.\ validation, 
% (stratified based on labels where appropriate), 
individually fits
various models, and finally creates an optimized model ensemble that
outperforms any of the individual trained models.  % on the validation data.  
For users willing to tolerate longer training times to
maximize predictive accuracy, \texttt{fit()} provides
additional options that may be specified:
\begin{itemize*}
\vskip -2em
\item \texttt{hyperparameter\_tune = True} optimizes the
  hyperparameters of the individual models.
\item \texttt{auto\_stack = True} adaptively chooses a model
  ensembling strategy based on bootstrap aggregation (bagging) and
  (multi-layer) stacking.
\item \texttt{time\_limits} controls the runtime of \texttt{fit()}.
\item \texttt{eval\_metric} specifies the metric used to evaluate predictive performance.
\end{itemize*}
% This makes AutoGluon easily accessible to novice ML practictioners
% without the need to write much code.

All intermediate results are saved on disk. If a call was canceled, we can invoke \texttt{fit()} with the argument \texttt{continue\_training=True}  to resume training. 
  
Note that \texttt{TabularPrediction} is just one of many tasks in the overall AutoGluon\footnote{\website} framework, which also supports AutoML on text and image data. It offers a large range of features including hyperparameter turning, neural architecture search, and distributed training, whose description lies beyond the scope of this work. This paper solely concentrates on the AutoGluon-Tabular module, referred to as AutoGluon here for simplicity, noting that design choices for structured data tables are radically different than for images/text (c.f.\ transfer learning).

% The rest part of AutoGluon is left for future works. In the following, we may refer AutoGluon-Tabular as AutoGluon for simplicity.  

\subsection{Data Processing}

When left unspecified by the user, the type of prediction problem at
hand is first inferred by AutoGluon based on the types of values
present in the label column. Non-numeric string values indicate a
classification problem (with the number of classes equal to the number
of unique values observed in this column), whereas numeric values with
few repeats indicate a regression problem. %Whenever only two
% noting classification and regression estimands are equivalent in this
% case under standard loss functions (for $y \in \{0, 1\}$: ${{\Pr(y = 1
%     \mid x)} = \mathbb{E}[y \mid x]}$). 
This simple feature is just one example of the many AutoGluon optimizations that help users quickly translate raw data into accurate predictions. 

AutoGluon relies on two sequential stages of data processing:
model-agnostic preprocessing that transforms the inputs to all models,
and model-specific preprocessing that is only applied to a copy of the
data used to train a particular model. Model-agnostic preprocessing
begins by categorizing each feature numeric, categorical, text, or
date/time.  
Uncategorized columns are discarded from the data, comprised of non-numeric, non-repeating fields with presumably little predictive value (e.g.\ UserIDs).
We consider text features to be columns of mostly unique strings, which on average contain more
than 3 non-adjacent whitespace characters. The values of each text column are transformed into numeric
vectors of $n$-gram features (only retaining those $n$-grams with high overall occurrence in the text columns to reduce memory footprint).
Date/time features are also transformed into suitable numeric
values. A copy of the resulting set of numeric and categorical
features is subsequently passed to model-specific methods for further
tailored preprocessing. To deal with missing discrete variables, we create an
additional \texttt{Unknown} category rather than imputing them. This
also allows AutoGluon to handle previously unseen categories at
test-time. Note that often observations are not missing at random and we
want to preserve the evidence of absence (rather than the absence of
evidence). 

\vskip -1in
\subsection{Types of Models}

We use a bespoke set of models in a predefined order. This ensures
that reliably performant models such as random forests are trained prior to more expensive and less reliable models such as $k$-nearest neighbors. This strategy is critical when stringent time-limits are imposed on \texttt{fit()}. It helped auto-sklearn previously win
time-constrained AutoML competitions \cite{feurer2018practical}.  In
particular, we consider neural networks, LightGBM boosted trees \cite{ke2017lightgbm}, CatBoost boosted trees
\cite{dorogush2018catboost}, Random Forests, Extremely Randomized Trees, and $k$-Nearest Neighbors. We use scikit-learn implementations of the latter three models. 
% Discussing hyperparameters is too complicated, there are tons, the link to hyperparameter descriptions in source code is in Appendix A.  The default hyperparameters are used for these models except for neural network that will be discussed shortly. \mli{double check if it's correct}
Note that this list
is far smaller than the multitude of candidates considered by AutoML frameworks like TPOT, Auto-WEKA, and auto-sklearn.
% We found that adding further models such as XGBoost, (Generalized) Linear Models, and Support Vector Machines we not an effective use of a given computational budget. 
Nonetheless, AutoGluon is sufficiently modular that users may easily add their own bespoke models into
the set of models that AutoGluon automatically trains, tunes, and ensembles.

\subsection{Neural Network}
\label{sec:nn}

%%%%%% NN diagram %%%%%%%%%
\begin{figure}[tb]
\centering
\includegraphics[width=0.7\columnwidth]{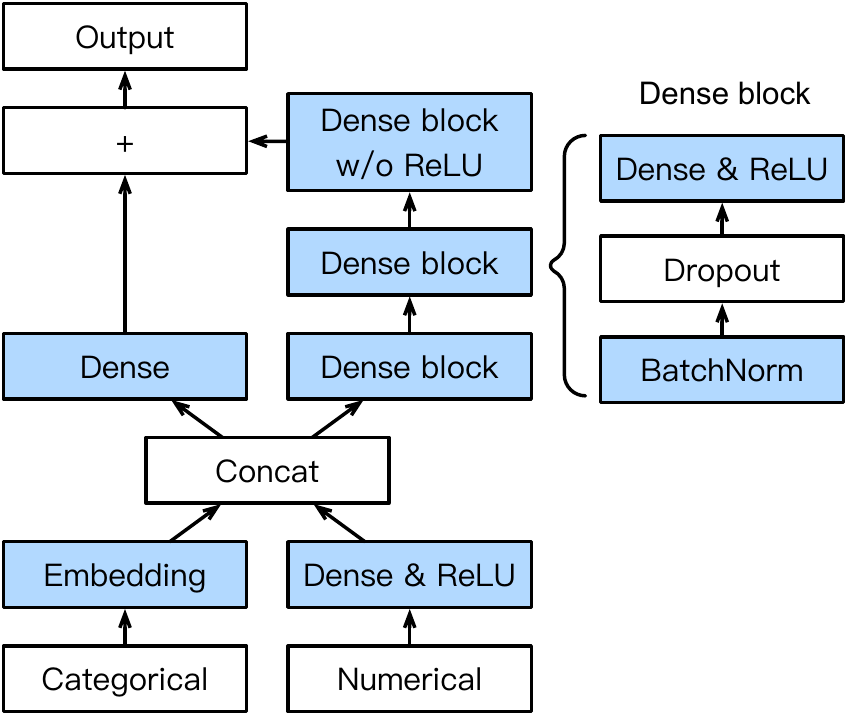}
\caption{Architecture of AutoGluon's neural network for tabular data composed of numerical and categorical features. Layers with learnable parameters are marked as blue.  % with 4 features (numeric and categorical features indicated by super-scripts $n$ and $c$). 
% Paths represent different transformations: embedding layer (red), dense layer (blue), rescaling by softmax/{$y$-range}  for classification/regression (orange), identity map (black).
% Red paths represent transformation by embedding layers, blue paths indicate dense layers, orange paths indicate rescaling (softmax for classification, inverse $(0,1) \rightarrow [y_{min}, y_{max}]$ map for regression), and black paths represent identity  transformations.
}
\label{fig:NN}
\end{figure}
%%%%%% End of NN diagram %%%%%%%%%

% Although deep learning can work for some (typically large) tabular
% datasets \cite{klambauer2017self}, there are many cases where tree
% ensembles reign supreme. Thus, most AutoML frameworks for structured data invested little
% effort on deep neural networks, often failing to utilize best practices in
% deep learning. For example, TPOT and auto-sklearn do not utilize
% multi-layer neural networks, Auto-WEKA uses a shallow feedforward
% network with sub-optimal sigmoid
% activations, and H2O employs standard MLPs, but trains them using
% Adadelta \cite{adadelta} which is often less stable than  other optimization
% algorithms, e.g.~\cite{adam}. 

Tabular data lacks the translation invariance and locality of images
and text that can be exploited via convolutions or recurrence.  
Instead, tabular datasets are comprised of
diverse types of values and thus feedforward networks are usually
the architecture of choice. However, the raw features in a data table
often already correspond to meaningful variables, better suited for
the axis-aligned single-variable splits of tree models, than a dense
feedforward layer that linearly blends all variables together into
individual hidden unit activation values.  Nonetheless,
\citet{mendoza2016towards} demonstrated that properly tuned neural
networks can provide significant accuracy-boosts when added to an
existing ensemble of other types of models. In particular, the
decision boundaries learned by neural networks differ from the
axis-aligned geometry of tree-based models, and thus provide valuable
diversity when ensembled with trees.

The network architecture used by AutoGluon is depicted in
Figure~\ref{fig:NN}, and additional details are in Appendix \ref{sec:details}.  It shares similar design choices as the models of
\citet{fastai,cheng2016wide}.  % Rather than preprocessing categorical features via one-hot encoding, 
Our network applies a separate
embedding layer to each categorical feature, where the
embedding dimension is selected proportionally to the number of unique
levels observed for this feature \cite{guo2016entity}. For
multivariate data, the individual embedding layers enable our network
to separately learn about each categorical feature before its representation is blended with other variables. % avoid comparing too much so our contributions are not overlooked. akin to a Factorization Machine \cite{rendle2010factorization}.
The embeddings of categorical features are concatenated with the
numerical features into a large vector which is both fed into a
3-layer feedforward network as well as directly connected to the
output predictions via a linear skip-connection. 

To our knowledge, AutoGluon is the first AutoML framework to use per-variable embeddings that are directly connected to the output via a linear shortcut path, which can improve their resulting quality via improved gradient flow. Most existing AutoML frameworks instead just apply standard feedforward architectures to one-hot encoded data \cite{kotthoff2017auto,H2O}.

\subsection{Multi-Layer Stack Ensembling}
\label{sec:ensembling}

Ensembles that combine predictions from multiple models have long
been known to outperform individual models, often
drastically reducing the variance of the final predictions
\cite{dietterich2000ensemble}.  All of the best-performing AutoML
frameworks today rely on some form of model ensembling such as bagging \cite{breiman1996bagging},
boosting \cite{freund1996experiments}, stacking
\cite{ting1997stacking}, or weighted combinations. 
In particular, various AutoML frameworks utilize shallow stack ensembling. Here a collection of individual ``base'' models are individually trained in the usual fashion. Subsequently, a ``stacker'' model is trained using the aggregated predictions of the base models as its features. The stacker model can improve upon shortcomings of the individual base predictions and exploit interactions between base models that offer enhanced predictive power \cite{van2007super}.

\begin{figure}[tb]
  \centering
  \includegraphics[width=0.75\columnwidth]{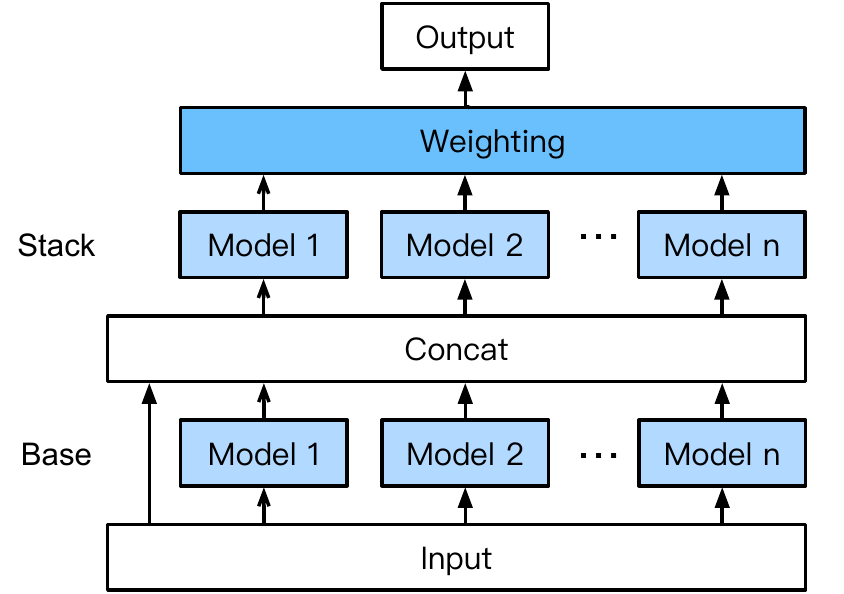}
\caption{AutoGluon's multi-layer stacking strategy, shown here using
  two stacking layers and $n$ types of base learners.} 
\label{fig:stacking}
\end{figure}

Multi-layer stacking feeds the predictions output by the stacker models as inputs to additional higher layer stacker models.
Iterating this process in multiple layers has been a winning strategy in prominent prediction competitions  \cite{netflix,TitericzSemenov16}. However, it is  nontrivial to implement robustly and thus not currently utilized by any AutoML framework.  
AutoGluon introduces a novel form of multi-layer stack ensembling, depicted in Figure~\ref{fig:stacking}.
Here the first layer has multiple base models, whose outputs are concatenated and then fed into the next layer, which itself consists of multiple stacker models. These stackers then act as base models to an additional layer.  

We extend the traditional stacking method with three changes that improve its resulting accuracy. 
To avoid another CASH problem, traditional stacking employs simpler models in the stacker than the base layers
\cite{van2007super}. AutoGluon instead simply reuses all of its base layer model types (with the same hyperparameter values) as stackers. 
This technique may be viewed as an alternative form of deep learning that utilizes layer-wise training, where the units connected between  layers may be arbitrary ML models.  
% a multi-path neural network such as Inception \cite{szegedy2015going}, but here the blocks consist of arbitrary ML models which are trained layer-wise.
Unlike existing strategies, our stacker models take as input not only the predictions of the models at the previous layer, but also the original data features themselves (input vectors are data features concatenated with lower-layer model predictions). 
Reminiscent of skip connections in deep learning, this enables our higher-layer stackers to revisit the original data values during training. %  which can improve the final result by exploiting predictive interactions between original features and lower-layer stacker predictions.  

Our final stacking layer applies  ensemble selection~\cite{caruana2004ensemble} to aggregate the stacker models' predictions in a weighted manner. 
AutoGluon's use of ensemble selection as the output layer of a stack ensemble is a strategy we have not previously encountered. 
While ensemble selection has been advocated for combining base models due to its resilience against  over-fitting  \cite{feurer2015efficient}, this property becomes even more valuable when aggregating predictions across a high-capacity stack of models.   % Figure~\ref{fig:stacking} illustrates the three changes. 

%%%%%% End of stacking diagram %%%%%%%%%

\subsection{Repeated $k$-fold Bagging}

AutoGluon further improves its stacking performance 
by utilizing all of the available data for both training and validation, through $k$-fold ensemble bagging of all models at all layers of the stack.  Also called cross-validated committees 
\cite{parmanto1996reducing}, $k$-fold bagging is a simple ensemble method that reduces  variance in the resulting predictions. 
This is achieved by randomly partitioning the data into $k$ disjoint chunks (we stratify based on labels), and subsequently training $k$ copies of a model with a different data chunk held-out from each copy. 
AutoGluon bags all models and each model is asked to produce out-of-fold (OOF) predictions on the chunk it did not see during training.  
As every training example is OOF for one of the bagged model copies,  this allows us to obtain OOF predictions from every model for every training example.

In stacking, it is critical that higher-layer models are only trained upon lower-layer OOF predictions. Training upon in-sample lower-layer predictions could amplify over-fitting and introduce covariate shift at test-time. 
Naive stacking with a traditional training/validation split in place of bagging can thus only use a fraction of the data to train the stacker. This issue becomes even more severe with multiple stacking layers.   
Our use of OOF predictions from bagged ensembles instead allows higher-layer stacker models to leverage the same amount of training data as those of the previous layer. 

While $k$-fold bagging efficiently reuses training data, it remains susceptible to a subtle form of over-fitting. The training process of certain models can be influenced by OOF data through factors such as the early stopping criterion, which can lead to minor over-fitting in OOF predictions. However, a stacker model trained on over-fit lower layer predictions may over-fit more aggressively, and OOF over-fitting can thus become amplified through each layer of the stack.

We propose a repeated bagging process to mitigate such over-fitting. When AutoGluon is given sufficient training time, it repeats the $k$-fold bagging process on $n$ different random partitions of the training data, averaging all OOF predictions over the repeated bags. The number of repetitions, $n$, is selected by estimating how many bagging rounds can be completed within the allotted training time.
OOF predictions which have been averaged across multiple $k$-fold bags display even less variance and are much less likely to be over-fit.
We find this $n$-repeated $k$-fold bagging process particularly  effective for smaller datasets where OOF over-fitting arises due to  limited OOF data sizes.

\subsection{Training Strategy} 
\label{sec:training}
Our overall training strategy is summarized in Algorithm~\ref{algo:ensemble}, where each stacking layer receives time budget $ T_\mathrm{total} / L$. 
In Step 7, AutoGluon first estimates the required training time and if this exceeds the remaining time for this layer, we skip to the next stacking layer. After each new model is trained, it is immediately saved to disk for fault tolerance. 
% Also note that all models in the same layer can be trained in parallel to accelerate the training. 
This design makes the framework highly predictable in its behavior:
both the time envelope and failure behavior are well-specified. This
approach guarantees that we can produce predictions as long as we can train at least one model on one fold within the allotted time. As we checkpoint intermediate iterations of sequentially trained models like neural networks and boosted/bagged trees, AutoGluon can still produce a model under meager time limits. We additionally anticipate that models may fail while training and skip to the next one in this event.

\begin{algorithm}[t!]
\begin{algorithmic}[1]
\REQUIRE data ($X$, $Y$), family of models $\mathcal{M}$, \# of layers $L$
  \STATE Preprocess data to extract features
  \FOR[Stacking]{$l = 1$ {\bfseries to} $L$}
     \FOR[$n$-repeated]{$i=1$ {\bfseries to} $n$}
         \STATE Randomly split data into $k$ chunks $\{X^j, Y^j\}_{j=1}^{k}$
         \FOR[$k$-fold bagging]{$j = 1$ {\bfseries to} $k$} 
             % \STATE Generate data fold $X^{-j}, Y^{-j}$ 
             \FOR{{\bfseries each} model type $m$ in $\mathcal{M}$}
                \STATE Train a type-$m$ model on $X^{-j}, Y^{-j}$
                \STATE Make predictions $\hat Y^{j}_{m,i}$ on OOF data $X^j$
             \ENDFOR
         \ENDFOR
     \ENDFOR
     \STATE Average OOF predictions $\hat Y_m=\{\frac1n\sum_{i}\hat Y^{j}_{m,i}\}_{j=1}^k$
     \STATE $X \gets$ concatenate$(X, \{\hat Y_m\}_{m\in\mathcal{M}})$  
  \ENDFOR
  % \WHILE{timeleft}
  % \ENDWHILE
\end{algorithmic}
\caption{AutoGluon-Tabular Training Strategy \\ (multi-layer stack ensembling + $n$-repeated $k$-fold bagging).}
\label{algo:ensemble}
\end{algorithm}

 Many AutoML frameworks train multiple models in parallel on the same
  instance. While this may save time in some cases, it leads to many
  out-of-memory errors on larger datasets without careful tuning. AutoGluon-Tabular instead trains models sequentially and relies on their individual implementations to efficiently leverage multiple cores. This allows us to train where other frameworks fail.

\begin{table*}[t!]
\vspace{-2mm}  
\centering
\caption{%
Popular AutoML frameworks for classification and regression with tabular
data. We indicate whether a table of {\textbf{\sc Raw}} data with
missing and non-numerical values can be handled automatically without
manual preprocessing and specification of feature types.}
\label{tab:overview}
\smallskip
\vspace{1mm}
\begin{small}
\begin{sc}
\begin{adjustbox}{max width=\textwidth}
\begin{tabular}{lcclll}
%\toprule
  \textbf{AutoML Framework} & \textbf{Open} & \textbf{Raw} &
  \textbf{Neural Network} & \textbf{CASH Strategy} & \textbf{Model Ensembling} \\
\midrule
Auto-WEKA & $\surd$ & $\times$ & sigmoid MLP & Bayesian Optimization & bag, boost, stack, vote \\[0.1em]
% & & & w/ sigmoid activation & & + stacking, voting \\[0.7em]
auto-sklearn & $\surd$ & $\times$ & none & BayesOpt + Meta-Learn & ensemble selection \\[0.1em]
% & & & & + meta-learning  & \\[0.7em]
TPOT & $\surd$ & $\times$ & none & genetic programming & stacking \\
% & & & & & \\
H2O & $\surd$ & $\surd$ & MLP + Adadelta & random search & stacking + bagging  \\[0.1em]
% & & & + Adadelta optimizer & & \\[0.7em]
GCP-Tables & $\times$ & $\surd$ & AdaNet (??) & AdaNet (??) & boosting (??)   \\[0.1em]
% & & & & & \\
% AutoPilot & & & & \\ % only method that does not do ensembling
AutoGluon   & $\surd$ & $\surd$ & embed categorical  &  fixed defaults  % $^\dagger$ 
& multi-layer stacking  \\[-0.1em]
 & & & + skip-connection & (set adaptively) & + repeated bagging 
% & & & & & + ensemble selection \\
%\bottomrule
\end{tabular}
\end{adjustbox}
\end{sc}
\end{small}
% \vspace{-2mm}  
\end{table*}

\section{AutoML Frameworks}
\label{sec:relwork}

% We found the following five are most relevant to AutoGluon-Tabular, which are summarized in \ref{tab:overview}.
Due to the immense potential of AutoML, many frameworks have been developed in this area.
We describe five widely used AutoML frameworks, summarized in Table \ref{tab:overview}.

%, even though these non-CASH aspects of the ML pipeline equally affect its resulting predictive performance. 

% \subsection{Auto-WEKA}
\textbf{Auto-WEKA} % , as developed by 
\cite{thornton2013auto} was one
of the first % comprehensive 
AutoML frameworks and remains popular
today with ongoing improvements \cite{kotthoff2017auto}. It relies on a wide array of models from the WEKA % \footnote{https://www.cs.waikato.ac.nz/ml/weka/} 
Java ML library, and is one of the first frameworks to 
% consider joint model and hyperparameter selection. This is addressed via Bayesian optimization over a conditional space of feasible solutions. 
address CASH via Bayesian optimization. After models have been selected, 
Auto-WEKA tries various ensembling strategies to further improve its predictions.
% Auto-WEKA uses Bayesian optimization to discover which models and hyperparameter values produce the best performance on a given dataset, treating this search as a single optimization problem over a highly-conditional space of feasible solutions. 
% While Auto-WEKA considers a large number of candidate models and assembles them into an ensemble with greater accuracy than any of the individual models, the only neural networks considered by this package are basic multilayer perceptrons with sigmoid activations. 

% \subsection{auto-sklearn}
\textbf{auto-sklearn}  % , as developed by 
\cite{feurer2015efficient} has been the winner of numerous AutoML competitions to date
\citep{feurer2018practical, feurer2019auto, guyon2019analysis}. 
This framework selects its base models from many options provided in the scikit-learn ML library \citep{scikit-learn}.
% This tool is built on top of the immensely popular scikit-learn ML library and selects base models from the plethora of options provided by it.
Two key factors driving auto-sklearn's success include its use of
meta-learning to warm start the hyperparameter search 
% on a new dataset based on hyperparameters that previously performed well on similar datasets 
\cite{feurer2014using}, as well as combining many models via the
ensemble selection strategy of \citet{caruana2004ensemble}.
Time management is a critical aspect of auto-sklearn, % \cite{feurer2018practical},
which leverages efficient multi-fidelity hyperparameter optimization strategies \cite{falkner2018bohb}.

% auto-sklearn carefully manages time \cite{feurer2018practical}, employing multi-fidelity hyperparameter optimization that is more time-efficient than traditional Bayesian optimization \cite{falkner2018bohb}.

% \subsection{TPOT}
\textbf{TPOT. }  The Tree-based Pipeline Optimization Tool (TPOT) of
\citet{olson2019tpot} employs genetic algorithms to optimize ML pipelines.  Each candidate consists of a choice data
processing operations, hyperparameters, models, and the option to stack ensembling with other 
models. % \cite{https://towardsdatascience.com/tpot-automated-machine-learning-in-python-4c063b3e5de9}.
While their evolutionary strategy can cope with this 
irregular search space, many of the randomly-assembled candidate
pipelines evaluated by TPOT end up invalid, % (with different stages lacking compatibility), 
thus, wasting valuable time that could be spent
training valid models.

% \subsection{H2O}
\textbf{H2O AutoML}  \citep{H2O} is perhaps the most widely used AutoML framework today, particularly in Kaggle prediction competitions. % within the Kaggle community that actively participates in prediction competitions where rapid ML prototyping is critical to success.  
% The most recent iteration, H2O-3, is an end-to-end AutoML framework, able to process raw CSV input and return predictions for test data.  
Able to process raw CSV input into predictions for test data, H2O employs one layer of ensemble stacking
combined with bagging, and utilizes an XGBoost tree ensemble \citep{chen2016} as one of
its strongest base models.  While it merely employs random search for
hyperparameter optimization, H2O frequently outperforms other AutoML frameworks
\cite{truong2019towards}.

% \subsection{GCP-Tables}
\textbf{GCP-Tables. } 
Recently released, Google Cloud Platform AutoML Tables is a commercial framework that handles end-to-end AutoML (raw data $\rightarrow$ predictions), but is only available on Google Cloud as a managed service  \cite{gcp}. 
% Released in 2019, it is currently in 'beta'. 
% GCP-Tables allows for end-to-end AutoML (raw data $\rightarrow$ predictions). 
GCP-Tables model training and predictions must be performed through
API calls to Google Cloud. The internals  of this framework thus remain unclear, although it is known to at least utilize both tree
ensembles and the AdaNet method of boosting neural network ensembles
\cite{cortes2017adanet}.  GCP-Tables is computationally
intensive, running a minimum of 92 instances in parallel when training
(\$19.32/hour with a minimum of 1 hour % of training time
regardless of dataset size).  
% GCP-Tables has produced impressive results on several Kaggle competitions, and is being integrated into Kaggle to serve as a baseline ML model for all future competitions \cite{gcpkaggle}.

% \subsection{Additional frameworks}

% Auto-XGBoost hasn't been updaetd in 10 months and has 86 stars on github, probably safe to not discuss in detail

\textbf{Other AutoML Platforms} worth mentioning  include:   auto-xgboost~\cite{autoxgboost}, GAMA~\cite{Gijsbers2019}, hyperopt-sklearn~\cite{Bergstra_2015}, TransmogrifAI~\cite{transmog}, ML-Plan~\cite{Mohr2018}, OBOE~\cite{chengrun2018oboe},
% including Auto-Keras, Darwin, AdaNet, and Amazon Sagemaker AutoPilot.
%Two recent AutoML libraries that focus on deep learning include 
 Auto-Keras~\cite{jin2019auto}, 
%is not evaluated here as lacking support for explicit time limits. In addition, 
as well as recent commercial AutoML solutions:  Sagemaker AutoPilot, %~\cite{autopilot}, \mli{citation formats are wrong. remove them for simplicity}
Azure ML, %~\cite{azureml}, 
H2O Driverless AI, % \cite{h20.ai}, 
DataRobot, %~\cite{datarobot}, 
and Darwin AutoML. %~\cite{darwin}.  

\begin{table*}[t!]
\centering
\caption{Comparing each AutoML framework against AutoGluon on the 39 AutoML Benchmark datasets (with 4h training time). 
Listed are the number of datasets where each framework produced: better predictions than AutoGluon (Wins), worse predictions (Losses), a system failure during training (Failures), or more accurate predictions than all of the other 5 frameworks (Champion). The latter 3 columns show the average: rank of the framework (among the 6 AutoML frameworks applied to each dataset), (rescaled) loss on the test data, and actual training time.  Averages are computed over only the subset of datasets/folds where all methods ran successfully. To ensure averaging over datasets remains meaningful, we rescale the loss values for each dataset such that they span $[0,1]$ among our AutoML frameworks. 
% performance  others is represented by column Best. Rank is the average ranked placement of each framework. 
% For each dataset, the framework with the lowest error gets a rank of 1 (1.5 in the case of a two way tie). Last place gets a rank of 6. 
% Rank and training time in minutes (Time) are averages across the competitions where \emph{every} framework ran successfully.
}
\label{tab:pair_openml_4h}
\vspace*{1em}

 \begin{footnotesize}
\begin{tabular}{lccccccc}
\toprule
\textbf{Framework} &  \textbf{Wins} &  \textbf{Losses} &  \textbf{Failures} &  \textbf{Champion} &  \textbf{Avg. Rank} &  \textbf{Avg. Rescaled Loss} & \textbf{Avg. Time (min)} \\
\midrule
         AutoGluon &              - &                - &                  \textbf{1} &                 \textbf{23} &              \textbf{1.8438} &                       \textbf{0.1385} &                      201 \\
        H2O AutoML &              4 &               26 &                  8 &                  2 &              3.1250 &                       0.2447 &                      220 \\
              TPOT &              6 &               27 &                  5 &                  5 &              3.3750 &                       0.2034 &                      235 \\
        GCP-Tables &              5 &               20 &                 14 &                  4 &              3.7500 &                       0.3336 &                      \textbf{195} \\
      auto-sklearn &              6 &               27 &                  6 &                  3 &              3.8125 &                       0.3197 &                      240 \\
         Auto-WEKA &              4 &               28 &                  6 &                  1 &              5.0938 &                       0.8001 &                      244 \\
\bottomrule
\end{tabular}

 \end{footnotesize}
% \end{center}
\vskip -0.1in
\end{table*}

\begin{table*}[t!]
\centering
\caption{Comparing each AutoML framework against AutoGluon on the 11 Kaggle competitions (under 4h time limit). 
Columns are defined as in Table \ref{tab:pair_openml_4h}. Instead of loss, we report the % (average) 
percentile rank, i.e.\ the proportion of teams beaten by AutoML on the competition leaderboard (higher is better).  
% Percentile Rank achieved on each competition's leaderboard (1 minus this value, so lower is better as in other results). 
Averages are computed only over the subset of 7 competitions where all methods ran successfully.
% Listed are the number of datasets where each framework produced: better predictions than AutoGluon ($>$ AutoGluon), worse predictions ($<$), equally accurate predictions ($=$), or a system failure during training (Failures).  
% Considering only the competitions where \emph{every}  framework ran successfully, the last 3 columns list each framework's average: percentile ranking on the competition leaderboards (1 - Percentile), number of minutes spent training (Time (m)), and the number of competitions where this framework outperformed all of the other five (Best).
% Averages are computed over only the subset of 7 competitions for which \emph{all} frameworks ran successfully (lower values are better).
}
\label{tab:pairkag4h}
\vspace*{1em}

 \begin{footnotesize}
\begin{tabular}{lccccccc}
\toprule
\textbf{Framework} &  \textbf{Wins} &  \textbf{Losses} &  \textbf{Failures} &  \textbf{Champion} &  \textbf{Avg. Rank} &  \textbf{Avg. Percentile} & \textbf{Avg. Time (min)} \\
\midrule
         AutoGluon &              - &                - &                  \textbf{0} &                  \textbf{7} &              \textbf{1.7143} &                    \textbf{0.7041} &                      \textbf{202} \\
        GCP-Tables &              3 &                7 &                  1 &                  3 &              2.2857 &                    0.6281 &                      222 \\
        H2O AutoML &              1 &                7 &                  3 &                  0 &              3.4286 &                    0.5129 &                      227 \\
              TPOT &              1 &                9 &                  1 &                  0 &              3.7143 &                    0.4711 &                      380 \\
      auto-sklearn &              3 &                8 &                  \textbf{0} &                  1 &              3.8571 &                    0.4819 &                      240 \\
         Auto-WEKA &              0 &               10 &                  1 &                  0 &              6.0000 &                    0.2056 &                      221 \\
\bottomrule
\end{tabular}

 \end{footnotesize}
% \end{center}
\vskip -0.1in
\end{table*}
% and other fully managed solutions offered by major cloud computing providers. 
% Darwin AutoML \cite{??} is a managed end-to-end AutoML service by the company sparkcognition.
% Amazon's Sagemaker AutoPilot \cite{??} is a recently released managed end-to-end AutoML service by Amazon.
% Due to the complexity of getting each of these AutoML frameworks working as intended and their varying levels of end-to-end AutoML support and open source accessibility, we have not evaluated them in our experiments. 
% Empirically comparing many AutoML tools, \citet{truong2019towards} found that H2O, auto-sklearn, and Auto-Keras generally outperformed the others. with H2O performing the best overall (no one tool dominated the others in their benchmark).

\section{Experiments}
\label{sec:exp}

\subsection{Setup}

% We will describe the benchmark datasets, experimental setups and present empirical results in this section. 
% \subsection{Benchmark Datasets}
AutoML platforms are nontrivial to compare as their relative performance differs between problems.  
% Previously adopted benchmarks like the UCI ML Repository % \cite{asuncion2007uci} 
% are mostly composed of older academic datasets with limited sample-sizes (including toy/synthetic data), and thus poorly reflect realistic applications facing ML practitioners today \citep{gijsbers2019open}. 
To ensure meaningful comparisons, we benchmark\footnote{Code to reproduce our benchmarks is available at: \\ \benchmarkgitlink} on a highly varied selection of 50 more curated datasets, spanning binary/multiclass classification and regression problems. These data are collected from two sources:

% We evaluate various AutoML frameworks on the following benchmarks\footnote{Code to reproduce all of our benchmarks is available at: \benchmarkgitlink}:

% \mli{use citation for URL to save space}.
\noindent\textbf{OpenML AutoML Benchmark. % ~\footnote{\url{https://openml.github.io/automlbenchmark/benchmark_datasets.html}}.  
}  39 datasets  curated by \citet{gijsbers2019open}  % that together comprise 
to serve as 
a representative benchmark for AutoML frameworks. These datasets span various binary and multi-class classification problems and exhibit substantial heterogeneity in sample size, dimensionality, and data types.   
Each AutoML framework is extensively evaluated in this benchmark through replicate runs on 10 different training/test splits for each dataset, making 390 prediction problems in total (splits provided by original benchmark, and reported numbers are averaged over them).  
We only provide the test data to AutoML frameworks at prediction time, and their predictions are scored using the original benchmark authors' code.  
As in the original benchmark, we train for both 1h as well as 4h,  and loss on each test set is calculated as $1 - \text{AUC}$ or log-loss for binary or multi-class classification tasks, respectively.
% In a previous evaluation of numerous AutoML frameworks using this data, H2O exhibited the highest accuracy by a slim margin \citep{gijsbers2019open}. 

\noindent\textbf{Kaggle Benchmark. }  11 tabular datasets chosen from recent Kaggle competitions to reflect real modern-day ML applications (full list in Table \ref{tab:kaggle}).  The competitions in this benchmark contain both regression and  (binary/multiclass) classification tasks. Various  metrics are used to evaluate predictive performance, each tailored to the particular applied problem by the competition organizers.  For every competition in this benchmark, none of the test data labels are available to us.  Each AutoML framework was trained on the provided training data, and then asked to make predictions on the provided (unlabeled) test data. These predictions were then submitted to Kaggle's server which evaluated their accuracy (on secret test labels) and provided a score 
% based on the performance metric specified in the competition
(details in Appendix~\S\ref{sec:kaggledets}). This benchmark offers a way to meaningfully compare AutoML performance across datasets: via the percentile rank achieved on the official competition leaderboards, which quantifies how many data science teams were outperformed by AutoML.   

% \subsection{Experimental Setup}

Using these datasets, we compared AutoGluon with the five aforementioned AutoML frameworks. Each was run with default settings except where the package authors suggested an improved setting in private communication.  
AutoGluon is run via the following code for every dataset: 

\vskip -0.0em
\begin{lstlisting}[language=Python]
task.fit(train_data, label, eval_metric,         
         time_limits=SEC, auto_stack=True)
\end{lstlisting}
\vskip -.5em

These settings instruct AutoGluon to utilize 2-layer stacking with (repeated) bagging, optimizing its predictions with respect to the specified evaluation metric as much as possible within the given time limit.  
While AutoGluon does support various hyperparameter optimization strategies, we do not utilize its \texttt{hyperparameter\_tune = True} option in this work. % reviewer will ask why. 
This demonstrates for the first time that high-accuracy AutoML is achievable entirely without CASH.   

All frameworks were run on the same type of EC2 cloud instance % machine % at AWS, 
with an identical training time limit, except GCP-Tables, which uses 92 Google Cloud servers by default. As some frameworks only loosely respected the specified time limits, we report actual training times as well.

%GCP refused to handle datasets with under 1000 rows, or over 1000 columns.

\subsection{Results}
\label{sec:results}

\begin{figure*}[tb] \centering
\hspace*{5mm} \includegraphics[width=0.55\textwidth]{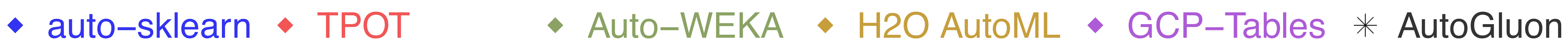} \\
\begin{tabular}{cc}
    \includegraphics[width=0.49\textwidth]{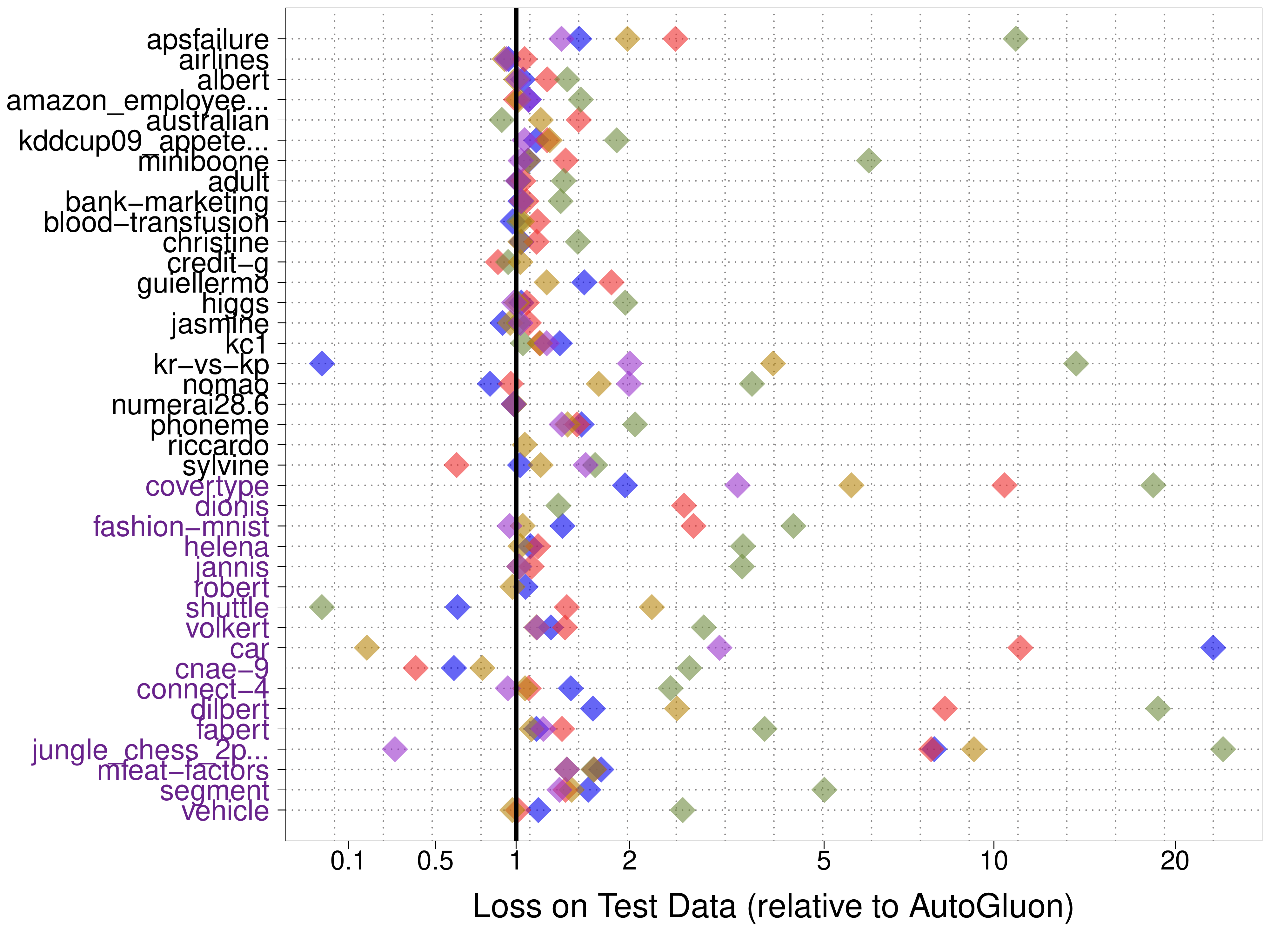}
&
    \includegraphics[width=0.49\textwidth]{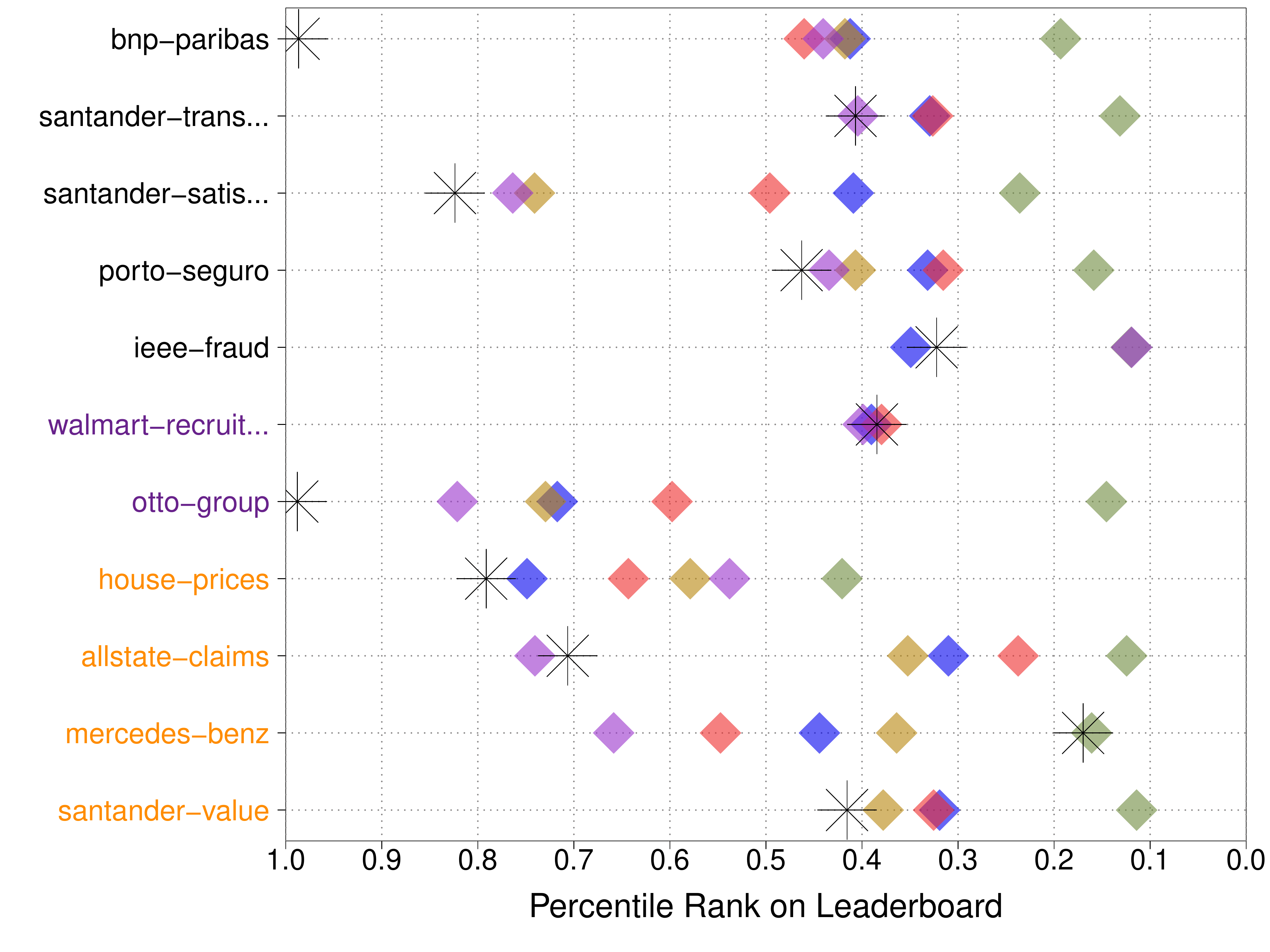} 
    \\
  \hspace*{20mm}  \textbf{(A)} AutoML Benchmark (1h) 
  & 
  \hspace*{20mm}  \textbf{(B)} Kaggle Benchmark (4h)
\end{tabular}
    \caption{
    \textbf{(A)} Performance of AutoML frameworks relative to AutoGluon on the AutoML Benchmark (with 1h training time).  
    \textbf{(B)}  Proportion of teams in each Kaggle competition whose scores were beat by each AutoML framework (with 4h training time).   
    Failed runs are not shown in these plots (and we omit a  massive loss of Auto-WEKA on \texttt{cars} as an outlier). The color of each dataset name indicates the task: binary classification (black), multi-class classification (purple), regression (orange).
    }
    \label{fig:benchmarkperf}
\end{figure*}

Tables \ref{tab:pair_openml_4h}-\ref{tab:pairkag4h} provide pairwise comparisons showing how each framework fares against AutoGluon in these benchmarks (after 4 hours of training), indicating how \emph{often} one framework is better than another.
Figure \ref{fig:benchmarkperf} depicts the performance of each framework across the many datasets in our benchmarks, indicating how \emph{much} better each framework is than the others for a particular problem.
Analogous results with other time limits (1 hour and 8 hours) can be found in the Appendix (Tables \ref{tab:pair_openml_1h}, \ref{tab:pairkag8h}, \ref{tab:rawlossopenml1h}-\ref{tab:percentileskagg8h} and Figures \ref{fig:openmlperf} and \ref{fig:kaggleperf}).

In any of these results, it is evident that AutoGluon is significantly more accurate than all of the other AutoML frameworks. For every benchmark and every time limit, AutoGluon is the only framework to rank better than 2\textsuperscript{nd} on average, indicating no other framework could beat it consistently.  
% This holds across all of the considered time limits.  
On over half of the datasets in each benchmark (23/39 for AutoML, 7/11 for Kaggle), AutoGluon performed better than \emph{all} of the other frameworks combined.  
AutoGluon is additionally more robust (with far less failures) and better at adhering to the specified training time limits (Table \ref{tab:pair_openml_4h}-\ref{tab:pairkag4h}, Figures \ref{fig:openmltime}-\ref{fig:kaggletime}).  
Beyond the benchmark-specific metrics for scoring predictions, AutoGluon also achieves the highest raw accuracy among AutoML frameworks when directly predicting class labels rather than probabilities (Table \ref{tab:pair_openml_accuracy_1h}).

\begin{table}[bt]
\centering
\caption{Performance of AutoML frameworks after 1h training vs.\ 4h training on each of the 39 AutoML Benchmark datasets. We count how many times the 1h variant performs better ($>$), worse ($<$), or comparably ($=$) to the 4h variant. 
}
\label{tab:pair_openml_1h_vs_4h}
\vspace*{1em}
 \begin{footnotesize}
\begin{tabular}{lccc}
\toprule
   \textbf{System} &  \textbf{$>$ 4h} &  \textbf{$<$ 4h} &  \textbf{$=$ 4h} \\
\midrule
    AutoGluon (1h) &                \textbf{5} &               \textbf{30} &                3 \\
   GCP-Tables (1h) &                8 &               16 &                1 \\
   H2O AutoML (1h) &                6 &               16 &                9 \\
 auto-sklearn (1h) &               12 &               16 &                1 \\
         TPOT (1h) &                6 &               24 &                2 \\
    Auto-WEKA (1h) &                7 &               16 &               10 \\
\bottomrule
\end{tabular}

 \end{footnotesize}
% \end{center}
\vskip -0.1in
\end{table}

AutoGluon's performance continues to improve with additional training time, and does so more reliably than the other frameworks which may start over-fitting (Table \ref{tab:pair_openml_1h_vs_4h}). 
% \ref{tab:pair_kaggle_4h8h24h}).  
When allowed to train for 24h on the \texttt{otto-group-product} data (with the same default arguments used in our benchmarks), AutoGluon was able to produce even more accurate predictions that achieved a rank of 23\textsuperscript{rd} place (out of 3505 teams) on the official leaderboard for this Kaggle competition. In just 24 hours and without any effort on our part, AutoGluon managed to outperform 99.3\% of the participating data scientists.
Even just the 4h training used in our benchmark  sufficed for AutoGluon to perform very well in some  competitions, placing 
42 / 3505 and 39 / 2920 on the official leaderboards of the \texttt{otto} and \texttt{bnp-paribas} competitions, respectively.

\subsection{Ablation Studies}
\label{sec:ablation}

Finally, we study the importance of AutoGluon's various components via ablation analysis. %  to determine how much each piece contributes to the overall framework's performance.
We run variants of AutoGluon with the following functionalities sequentially removed: First, we omit iterated  repetitions of bagging, just using a single round of $k$-fold  bagging (\emph{NoRepeat}). Second, we omit our multi-layer stacking strategy, so the resulting model ensemble only uses bagging and ensemble selection (\emph{NoMultiStack}). Third, we omit bagging and rely on ensemble selection with only a single training/validation split of the data 
% , where ensemble selection of base models is guided by validation performance 
% over this validation data subset 
(\emph{NoBag}). 
%  where ensemble selection is applied based on only  a small heldout subset of the training data (just a single training/validation split). 
Fourth, we omit our neural network from the set of base models that are applied (\emph{NoNetwork}). 
% Lastly, we omit ensembling altogether, performing inference with whichever individual base model exhibited the highest validation score (\emph{NoEnsemble}).  

% Results on the AutoML benchmark are shown in 
Table \ref{tab:ablation_4h_mini} shows  that 
% As expected, the 
overall predictive performance dropped each time we removed the next feature in this list. 
% With 4h training time, we observe the following average (rescaled) loss values: \emph{Full AutoGluon} = 0.14, \emph{NoRepeat} = 0.21,  \emph{NoMultiStack} = 0.44, \emph{NoBag} = 0.66, \emph{NoNetwork} = 0.74. 
Thus, these features all entail key reasons why AutoGluon is more accurate than other AutoML frameworks.
% The most pronounced drops accompanied the removal of stacking and our neural network, identifying these multi-layered components as key reasons why AutoGluon is more accurate than the other AutoML frameworks.  
Even after 4h of training, the  \emph{NoMultiStack} variant could usually not outperform the full version of AutoGluon trained for only 1h.

% Even though our neural network is often not the most accurate base model, it contributes valuable diversity to the ensemble pool and is thus often upweighted during ensemble selection. 

% It is important to assess the contributions of each individual AutoGluon component in the context of the other pieces (for example, certain base models may tend to ensemble better with others even though their individual predictive performance is not as strong).  

\begin{table}[t!] 
\centering
\vspace*{-3mm}
\caption{
Ablation study of AutoGluon on the AutoML Benchmark (4h training time). Columns are defined as in Table \ref{tab:pair_openml_4h}.
% To ensure averaging over different datasets remains meaningful, we rescale the loss values for each dataset such that they span $[0,1]$ among our AutoML frameworks. The rescaled loss for a dataset is set = 0 for the champion variant and = 1 for the worst-performing variant. The remaining frameworks are linearly scaled between these endpoints based on their relative loss.
}
\label{tab:ablation_4h_mini}
\vskip 0.1in
\begin{tabular}{lcc}
\toprule
\textbf{Framework} &  \textbf{Avg. Rank} &  \textbf{Avg. Rescaled Loss} \\
\midrule
         AutoGluon &              \textbf{1.9324} &                       \textbf{0.1660} \\
          NoRepeat &              2.1216 &                       0.2199 \\
      NoMultiStack &              2.8514 &                       0.5237 \\
             NoBag &              3.9054 &                       0.7199 \\
         NoNetwork &              4.1892 &                       0.8171 \\
\bottomrule
\end{tabular}

\vskip 0.1in
\end{table}

\section{Conclusion}
\label{sec:disc}

This paper introduced AutoGluon-Tabular, an AutoML framework for structured data that automatically manages the end-to-end ML pipeline. AutoGluon codifies best modeling practices from the data science community, and extends them in various ways. Key aspects of AutoGluon-Tabular include its robust data processing to handle heterogeneous datasets, modern neural network architecture, and powerful model ensembling based on novel combinations of multi-layer stacking and repeated $k$-fold bagging.  Our comprehensive empirical evaluation reveals that AutoGluon-Tabular is significantly more accurate than popular AutoML frameworks that focus on combined algorithm selection and hyperparameter optimization (CASH). Although AutoML is often taken as synonymous with CASH, our work clearly demonstrates that it only constitutes one piece of a successful end-to-end AutoML framework.

\clearpage
\balance
\bibliography{autogluon-automl}
\bibliographystyle{icml2020}

%% Supplement %%
%% Supplement %%

%%%%%%%%%%%%%%%%%%%%%%%%%%%%%%%%%%%%%%%%%%%%%%%%
%%% Supplement %%%%
%%%%%%%%%%%%%%%%%%%%%%%%%%%%%%%%%%%%%%%%%%%%%%%%
%   \endinput % hide appendix for submission

\clearpage \newpage
\beginsupplement
\onecolumn
\appendix

\def\toptitlebar{\hrule height1pt \vskip .2in} 
\def\bottomtitlebar{\vskip .22in \hrule height1pt \vskip .3in} 

\thispagestyle{plain}

\setcounter{page}{1}
\pagenumbering{arabic}
\setlength{\footskip}{20pt}  
\vspace*{-3.5mm}
\begin{center}
\toptitlebar
{\Large \bf Appendix: \\[0.2em]
\papertitle 
}
\bottomtitlebar
\end{center}

\FloatBarrier

% \tableofcontents

\section{AutoGluon Implementation Details} 
\label{sec:details}

Default hyperparameter values used for each model can be found in the AutoGluon source code: \gitlink{}.
The implementation of each model and its hyperparameter values are located in the directory: 
\texttt{autogluon/utils/tabular/ml/models/}. 
These default hyperparameter values were chosen a priori and not tuned based on the particular datasets used for benchmarking in this paper. 

Implemented using MXNet Gluon, the neural network in AutoGluon employs ReLU activations, dropout regularization, batch normalization, Adam with a weight decay penalty \citesi{adam}, and early stopping based on validation performance. While not tuned via hyperparameter optimization, the size of the hidden layers is nonetheless scaled adaptively based on properties of the training data (in a fixed manner). In particular, the feedforward branch of our model uses hidden layers of size 256 and 128 which are additionally  scaled up based on the number of classes in multi-class settings.  The width of the numeric embedding layer ranges between 32 - 2056 and is determined based on the number of numeric features and the proportion of numeric vs.\ categorical features. Inspired by \citetsi{fastai}, a discrete feature with $k$ unique categories observed in the training data is processed by an embedding layer of size: $1.6 \times k^{0.56}$ (up to threshold of size 100).

To ensure stable gradients in regression we rescale targets and use the $L_1$ loss (even for mean-squared-error objectives). 
% While recent work found that cyclic learning rate schedules greatly improve convolutional networks in computer vision, we did not witness the same benefits in our applications with tabular data. Instead, our neural network weights are simply trained via the standard Adam optimizer. In such applications, we find it is rather more important to consider when to stop updating the weights, as over-fitting is a common issue in the absence of early stopping.
While SeLU activations have demonstrated strong performance in tabular data applications \citesi{klambauer2017self}, we found them occasionally unreliable on data with peculiar characteristics, and opted to use simpler ReLU units for the sake of robustness. 

Model-specific data preprocessing  for our neural network included the following steps. For numeric features: missing values were imputed using the median, 
quantile normalization was applied to 
skewed distributions, and mean-zero unit-variance rescaling was applied to all other variables.
For categorical features: a separate ``Unknown'' category
was introduced for missing data as well as new categories encountered during inference, and an ``Other'' category was used to handle high-cardinality features with $> 100$ possible levels, where all rare categories were reassigned to ``Other''. 
We only applied our embedding layers to categorical features with at least 4 discrete levels (others were simply one-hot encoded and then treated as numeric).

\iffalse
We highlight additional issues encountered during fully automatic data preprocessing, in order to provide a taste of the subtle complexities that can arise.
Consider for example a categorical feature $X$ which takes thousands of discrete levels, has missing value in some training examples, and takes new values in the test data that were never observed during training. For such a feature, AutoGluon first creates an extra \texttt{Unknown} category to which unknown values encountered during inference can be mapped, so that they can be handled by the existing models. Missing categorical values are also assigned to this \texttt{Unknown} category rather than being imputed, in order to preserve potentially predictive information in the missingness-pattern itself (imputation methods in contrast heavily rely on a missing at random assumption that is often violated \citesi{missingatrandomfalse??}). To avoid having to learn about the thousands of different categories corresponding to $X$ (which is statistically infeasible given limited data), AutoGluon only keeps the top categories with the highest rate of occurrence in the training data; all other values of $X$ are mapped to a single \texttt{Other} category.  
However, if $X$ takes too many possible values, for example a different unique value in each training example (e.g.\ name/address), then AutoGluon will automatically discard this feature from the set of predictors under consideration.
\fi

\section{Data used in Kaggle Benchmark}
\label{sec:kaggledets}

\begin{table*}[!bth]
% \begin{center}
\centering
\caption{ % Overview of 
Summary of the 11 Kaggle competitions used in our benchmark, including: the date of each competition, around how many teams participated, and the number of rows/columns in the provided training data.  
The metrics used to evaluate predictive performance in each competition include: root mean squared logarithmic error (RMSLE), coefficient of determination ($R^2$), mean absolute error (MAE), logarithmic loss (log-loss), 
area under the Receiver Operating Characteristic curve (AUC), and normalized Gini index (Gini).  
}
\label{tab:kaggle}
\vskip 0.15in

% \begin{small}
\begin{tabular}{llllllll}
\toprule
\textbf{Competition} & \textbf{Task} & \textbf{Metric} & \textbf{Year} & \textbf{Teams} & \textbf{Rows} &  \textbf{Colums}  \\
\midrule
house-prices-advanced-regression-techniques & regression & RMSLE & current & 5100 & 1460 & 80  \\
mercedes-benz-greener-manufacturing & regression & $R^2$ & 2017 & 3800 & 4209 & 377 \\
 santander-value-prediction-challenge & regression & RMSLE & 2019 & 4500 & 4459 & 4992  \\
allstate-claims-severity & regression & MAE & 2017 & 3000 & 1.8E+5 & 131 \\
bnp-paribas-cardif-claims-management & binary & log-loss & 2016  & 2900 & 1.1E+5 & 132 \\ 
santander-customer-transaction-prediction & binary & AUC & 2019 & 8800 & 2.2E+5 & 201 \\
santander-customer-satisfaction & binary & AUC & 2016 & 5100 & 7.6E+4 & 370 \\
porto-seguro-safe-driver-prediction & binary & Gini & 2018 & 5200 & 6.0E+5 & 58 \\
ieee-fraud-detection & binary & AUC & 2019 & 6400 & 5.9E+5 & 432 \\
walmart-recruiting-trip-type-classification & multi-class & log-loss & 2016 & 1000 & 6.5E+5 & 7  \\
otto-group-product-classification-challenge & multi-class & log-loss & 2015 & 3500 & 6.2E+4 & 94\\

\bottomrule
\end{tabular}
% \end{small}
% \end{center}
\vskip 5pt
\end{table*}

Table \ref{tab:kaggle} describes the datasets that comprised our Kaggle benchmark.  Data for each competition can be obtained from its website: \url{kaggle.com/c/x/} where \url{x} is the name of the competition specified in the table.  
We selected data for the benchmark based on a few criteria.  
First, we aimed to include datasets for which \citetsi{gcp, gcpkaggle} previously demonstrated that GCP-Tables could produce strong results (indicating these are suitable  candidates for AutoML). 
These competitions included the following: \texttt{allstate-claims-severity}, \texttt{porto-seguro-safe-driver-prediction}, 
\texttt{walmart-recruiting-trip-type-classification},
\texttt{ieee-fraud-detection}.  
We decided not to include the other three datasets from \citetsi{gcp} in our benchmark because either the data are unavailable (\texttt{criteo-display-ad-challenge}), the 
competition is no longer scoring predictions (\texttt{mercari-price-suggestion-challenge}), or the data require  manual transformations to be formatted as a single table, without a clear canonical recipe to construct such a table  
(\texttt{kdd-cup-2014-predicting-excitement-at-donors-choose}).

The remaining benchmark data were selected by optimizing for a mix of regression and binary/multiclass classification tasks with IID data (i.e.\ without temporal dependence), while favoring competitions that were either more recent (indicating more timely applications) or had a large number of teams competing (indicating more prominent applications). Here, we chose to disregard competitions that provide multiple data files needing to be manually joined to obtain a single data table (except for \texttt{ieee-fraud-detection} from \citetsi{gcpkaggle} which had an obvious join strategy).  Typically based on domain-specific knowledge, the precise manner in which such manual joins are conducted will heavily affect predictive performance, and lies beyond the scope of current AutoML frameworks (which assume the data are contained within a single table).  
Beyond formatting them into a single table as necessary and specifying ID columns when they are needed to submit predictions, the data provided by Kaggle were otherwise not altered (no feature selection/engineering), in order to evaluate how our AutoML frameworks perform on raw data.  
For posterity, Table \ref{tab:badkaggle} lists additional competitions that appear to fit our selection criteria, but were ill-suited for the benchmark upon closer inspection.  

Predictions submitted to a Kaggle competition receive two scores evaluated on private and public subsets of the test data.  
The performance reported in this paper is based on the private score from each Kaggle competition, which is used to decide the official leaderboard. An exception is the currently ongoing  \texttt{house-prices-advanced-regression-techniques} competition for which we report public scores, as the private scores are not yet available (The leaderboard ranks achieved by our AutoML frameworks on this competition are thus unreliable as many competitors game the public scores through various exploits; public test scores nonetheless suffice for fair comparison of AutoML frameworks since our frameworks solely access the test data to make predictions). 
Throughout our presented results, two methods' performance is deemed equal if their predictions were scored within 5 decimal places of each other. 

Certain competitions used scoring metrics not supported by some of the AutoML frameworks in our evaluations. However, by suitably processing the data, we were able to ensure every AutoML framework  optimizes a metric monotonically related to the true scoring function.
For example, we log-transformed the $y$-values from competitions using the Root Mean Squared Logarithmic Error, then specified that each AutoML frameworks should use the mean-squared error metric, and finally applied the inverse transformation to their predictions before submitting them to Kaggle.  Normalized Gini-index scoring was handled by instead specifying the proportional AUC metric to each AutoML framework.  
 Thus, each AutoML tool was informed of the exact evaluation metric that would be employed and our comparisons are fully equitable.

% Table of bad datasets:
\begin{table*}[!bth] 
% \begin{center}
\centering
\caption{ Other prominent tabular datasets from Kaggle that are \emph{not} suited for existing AutoML tools (not included in our benchmark).
}
\label{tab:badkaggle}
\vskip 0.15in

% \begin{small}
\begin{tabular}{lr}
\toprule
\textbf{Competition} & \textbf{Not appropriate for AutoML because...}   \\
\midrule
ga-customer-revenue-prediction & unsuited for ML without extensive manual preprocessing
\\
microsoft-malware-prediction & non IID data with temporal dependence, shift in test distribution
\\ 
talkingdata-adtracking-fraud-detection & non IID data with temporal dependence
\\
bigquery-geotab-intersection-congestion & peculiar prediction problem, non IID data, needs special preprocessing
\\
elo-merchant-category-recommendation & requires manual join of multiple data files
\\
restaurant-revenue-prediction & minute sample size ($n=137$) 
\\
new-york-city-taxi-fare-prediction & geospatial data requiring domain knowledge or external information sources
\\
nyc-taxi-trip-duration & geospatial data requiring domain knowledge or external information sources
\\
caterpillar-tube-pricing & requires manual join of multiple data files
\\
favorita-grocery-sales-forecasting & non IID data, multiple data files that require joining 
\\
walmart-recruiting-sales-in-stormy-weather & requires manual join of multiple data files
\\
rossmann-store-sales & non IID data with temporal dependence
\\
bike-sharing-demand & non IID data with temporal dependence 
\\
LANL-Earthquake-Prediction & non IID data with temporal dependence \\
ashrae-energy-prediction & non IID data with temporal dependence 
\\
\bottomrule
\end{tabular}
% \end{small}
% \end{center}
\vskip 0.1in
\end{table*} % Bad datasets

\FloatBarrier
\section{Details Regarding Usage of AutoML frameworks}
\label{sec:baselinedetails}

Code to reproduce our benchmarks is available here: \benchmarkgitlink{}.  
Our evaluations are based on running each AutoML system on every dataset in the exact same manner.
Having to manually adjust tools to particular datasets would otherwise undermine the purpose of automated machine learning. 

Only H2O and GCP-Tables could robustly handle training with CSV files of raw data in our experiments (each utilizing their own automated inference of feature types).   While Auto-WEKA aims to do the same, our experiments produced numerous errors when applying Auto-WEKA to raw data (e.g.\ when a new feature-category appeared in test data).  To enhance its robustness, we provided Auto-WEKA with the same preprocessed data that we provided to TPOT and auto-sklearn. Lacking end-to-end AutoML capabilities, these packages do not support raw data input and require the data to be preprocessed. % (missing values imputed, categorical features encoded into numeric, etc.). 
Thus, we provided Auto-WEKA, TPOT, and auto-sklearn with the same preprocessed version of each dataset, producing via the same steps AutoGluon uses to transform raw data into numerical features that are fed to certain models:  Inferred categorical features are restricted to only their top 100 categories, then one-hot encoded into a vector representation with additional categories to represent rare categories, missing values, and new  categories only encountered at inference-time. Inferred numerical features have their missing values imputed and then are rescaled to zero mean and unit variance.  

We find that given the AutoGluon-processed data, Auto-WEKA, TPOT, and auto-sklearn are able to match their performance in the original AutoML benchmark (Table \ref{tab:pair_openml_new_vs_old}), this time without requiring that the feature types have been manually specified for each package.
% TPOT, auto-sklearn, and Auto-WEKA required alterations to serve as fully end-to-end AutoML frameworks for our benchmark. These frameworks do not support arbitrary data input such as CSV. 
%The data must be preprocessed to have missing values imputed, categorical features encoded into numeric, and data normalized prior to being fit. To enable their inclusion in the benchmark, we first preprocess the data through AutoGluon's preprocessing module. For inferred categorical features, we one-hot encode the most frequent categories, and create an additional category to represent rare categories. This is roughly equivalent to the preprocessing done in the AutoML Benchmark for these frameworks, and achieves similar performance. 
As a commercial cloud service, GCP-Tables differed from the other tools in that it is fully automated with no user-specified parameters to affect predictive performance
beyond the given evaluation metric and time limits. % (which were only loosely respected in our experiments).  

Where available, we used newer versions of each open-source AutoML framework than those \citetsi{gijsbers2019open} evaluated in the original AutoML Benchmark.  In particular, we used TPOT version 0.11.1, Auto-WEKA 2.6, H2O 3.28.0.1, and auto-sklearn version\footnote{While a  newer 0.6.0 auto-sklearn version exists, it has \texttt{sckit-learn} dependency that is incompatible with AutoGluon preventing them from being installed together. There does not appear to be any updates to auto-sklearn's ML/modeling process between versions 0.5.2 and 0.6.0} 0.5.2.   
For each of these AutoML libraries, we confirmed with the original package authors that any modifications we made to the default AutoML benchmark settings would be improvements.  
The code to run each AutoML tool is found in the  \texttt{autogluon\_utils/benchmarking/baselines/} folder of our linked code.  
Because running GCP-Tables on all 390 prediction problems of the AutoML benchmark was economically infeasible (4h runs would total over \$30k), we only ran this AutoML tool on the first fold of each of the 39 datasets. All pairwise comparisons between GCP-Tables and AutoGluon only consider the AutoGluon performance over this same fold, rather than all 10.

We followed the protocol of the original AutoML Benchmark and trained frameworks with 1h and 4h time limits.  The Kaggle datasets tend to be larger than those of the AutoML Benchmark and posed memory issues for some of the baseline AutoML tools.  To ensure no AutoML framework is resource-limited, we ran the Kaggle benchmark for longer than the AutoML datasets (4h and 8h time limits), and used more powerful AWS \texttt{m5.24xlarge} EC2 instances (384 GiB memory, 96 vCPU cores).  For the AutoML Benchmark, we used the same machine as in the original benchmark, an AWS \texttt{m5.2xlarge} EC2 instance (32 GiB memory, 8 vCPU cores).

To ensure averaging over different datasets remains meaningful in the AutoML Benchmark, we report loss values over the test data that have been rescaled. We rescale the loss values for each dataset such that they span $[0,1]$ among our AutoML frameworks. The rescaled loss for a dataset is set = 0 for the champion framework and = 1 for the worst-performing framework. The remaining frameworks are linearly scaled between these endpoints based on their relative loss. To ensure all head-to-head comparisons between frameworks remain fair, our reported averages/counts are taken \emph{only} over those datasets where all frameworks trained without error.

\begin{table*}[h!]
\centering
\caption{Comparing our usage of AutoML systems  against the results from the original AutoML Benchmark \citesi{gijsbers2019open}. Out of the 39 datasets, we count how often our implementation exceeded the original performance ($>$), or fell below the original performance ($<$), or was equally performant ($=$). Since there were no ties here,  all missing counts are datasets where one framework failed.
Rather than providing TPOT, auto-sklearn, and Auto-WEKA with information about the true feature types (as done in the original benchmark), we instead provided them with data automatically preprocessed by AutoGluon. This allows these methods to be applied in a more automated/robust manner to other datasets, without harming their performance.
}
\label{tab:pair_openml_new_vs_old}
\vspace*{1em}

 \begin{footnotesize}
\begin{tabular}{lccc}
\toprule
   \textbf{System} &  \textbf{$>$ Original} &  \textbf{$<$ Original} &  \textbf{$=$ Original} \\
\midrule
   H2O AutoML (1h) &                     18 &                     16 &                      0 \\
 auto-sklearn (1h) &                     16 &                     14 &                      0 \\
         TPOT (1h) &                     17 &                     13 &                      0 \\
    Auto-WEKA (1h) &                     18 &                     12 &                      0 \\
   H2O AutoML (4h) &                     15 &                     15 &                      0 \\
 auto-sklearn (4h) &                     15 &                     17 &                      0 \\
         TPOT (4h) &                     13 &                     16 &                      0 \\
    Auto-WEKA (4h) &                     17 &                     12 &                      0 \\
\bottomrule
\end{tabular}

 \end{footnotesize}
% \end{center}
\vskip 0.1in
\end{table*}

\subsection{AWS Sagemaker AutoPilot}
\label{sec:autopilot}

Like GCP-Tables, Sagemaker AutoPilot is another cloud AutoML service which allows users to automatically obtain predictions from raw data  with a single API call or just a few clicks. It runs a number of algorithms and tunes their hyperparameters on fully managed compute infrastructure, but does not utilize any model ensembling. 
% Currently, AutoPilot supports automatic data cleaning and preprocessing, automatic algorithm selection for regression, binary classification, and multi-class classification, automatic hyperparameter optimization, distributed training, and automatic instance and cluster size selection. 
At this time, AutoPilot does not estimate class probabilities (only produces class predictions). In order to receive a score in our benchmarks (log-loss, AUC, etc.), predicted class probabilities are needed.  For that reason, we only included AutoPilot in the raw accuracy comparison in Table \ref{tab:pair_openml_accuracy_1h}.

\FloatBarrier
\clearpage 
\section{Additional Results}
\label{sec:extraresults}

This section contains the comprehensive set of results from all of our benchmarks.  

\subsection{Additional Results for AutoML Benchmark}

\begin{table*}[hbt!]
\centering
\caption{Comparing each AutoML framework against AutoGluon on the 39 AutoML Benchmark datasets (with 1h training time). 
Listed are the number of datasets where each framework produced: better predictions than AutoGluon (Wins), worse predictions (Losses), a system failure during training (Failures), or more accurate predictions than all of the other 5 frameworks (Champion). The latter 3 columns show the average: rank of the framework (among the 6 AutoML tools applied to each dataset), (rescaled) loss on the test data, and actual training time.  Averages are computed over only the subset of datasets/folds where all methods ran successfully. 
 Recall that loss on each test set is evaluated as 1 - AUC or log-loss for binary or multi-class classification tasks, respectively (lower = better).
 To ensure averaging over different datasets remains meaningful, we rescale the loss values for each dataset such that they span $[0,1]$ among our AutoML frameworks. The rescaled loss for a dataset is set = 0 for the champion framework and = 1 for the worst-performing framework. The remaining frameworks are linearly scaled between these endpoints based on their relative loss.
}
\label{tab:pair_openml_1h}
\vspace*{1em}

 \begin{footnotesize}
\begin{tabular}{lccccccc}
\toprule
\textbf{Framework} &  \textbf{Wins} &  \textbf{Losses} &  \textbf{Failures} &  \textbf{Champion} &  \textbf{Avg. Rank} &  \textbf{Avg. Rescaled Loss} & \textbf{Avg. Time (min)} \\
\midrule
         AutoGluon &              0 &                0 &                  0 &                 19 &              1.5455 &                       0.0474 &                       57 \\
        GCP-Tables &              6 &               20 &                 13 &                  5 &              2.8182 &                       0.2010 &                       90 \\
        H2O AutoML &              8 &               30 &                  1 &                  5 &              3.1818 &                       0.1914 &                       58 \\
      auto-sklearn &              8 &               26 &                  5 &                  4 &              3.7273 &                       0.2176 &                       60 \\
              TPOT &              5 &               30 &                  4 &                  4 &              4.0909 &                       0.2900 &                       67 \\
         Auto-WEKA &              4 &               31 &                  4 &                  2 &              5.6364 &                       0.9383 &                       62 \\
\bottomrule
\end{tabular}

 \end{footnotesize}
% \end{center}
\vskip 0.1in
\end{table*}

\begin{figure*}[h!] \centering
\includegraphics[width=0.58\textwidth]{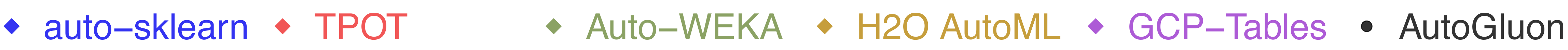} 
\\
    \includegraphics[width=0.5\textwidth]{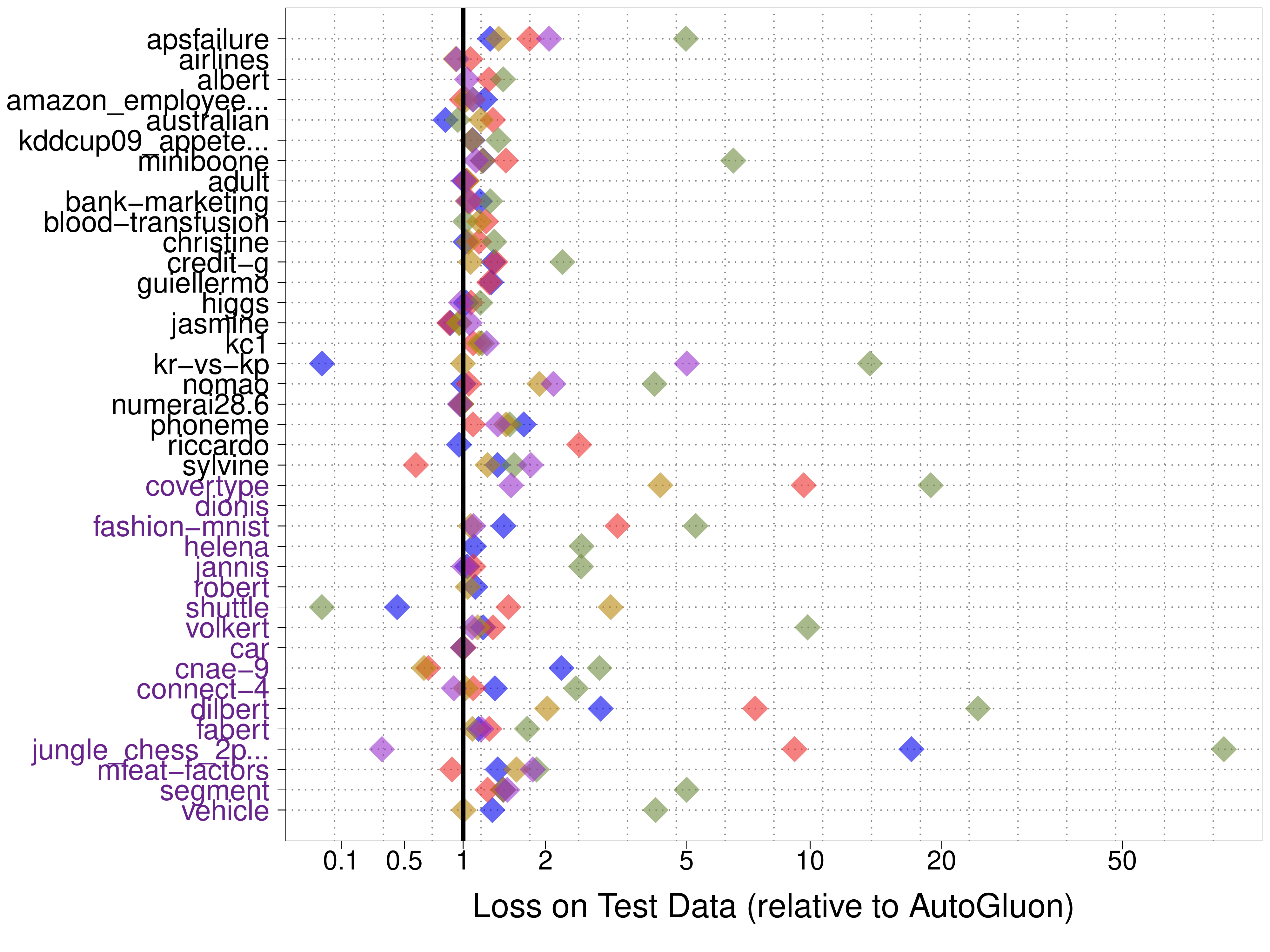}
    \caption{Performance of AutoML frameworks relative to AutoGluon on each dataset from the AutoML Benchmark (under 4h training time limit).
    Failed runs are not shown here (and we omit a  massive loss of Auto-WEKA on \texttt{cars} as an outlier). Loss is measured via $1 - \text{AUC}$ for binary classification datasets (black text), or log-loss for multi-class classification datasets (purple text), and is divided by AutoGluon's loss here.
    % on the cars dataset, where it was possible to achieve extremely low log-loss.
    }
    \label{fig:openmlperf}
\end{figure*}

% TRAINING TIMES OPENML:
\begin{figure*}[h!] \centering
\hspace*{5mm} \includegraphics[width=0.58\textwidth]{LegendPlot.pdf} \\
\begin{tabular}{cc}
    \includegraphics[width=0.49\textwidth]{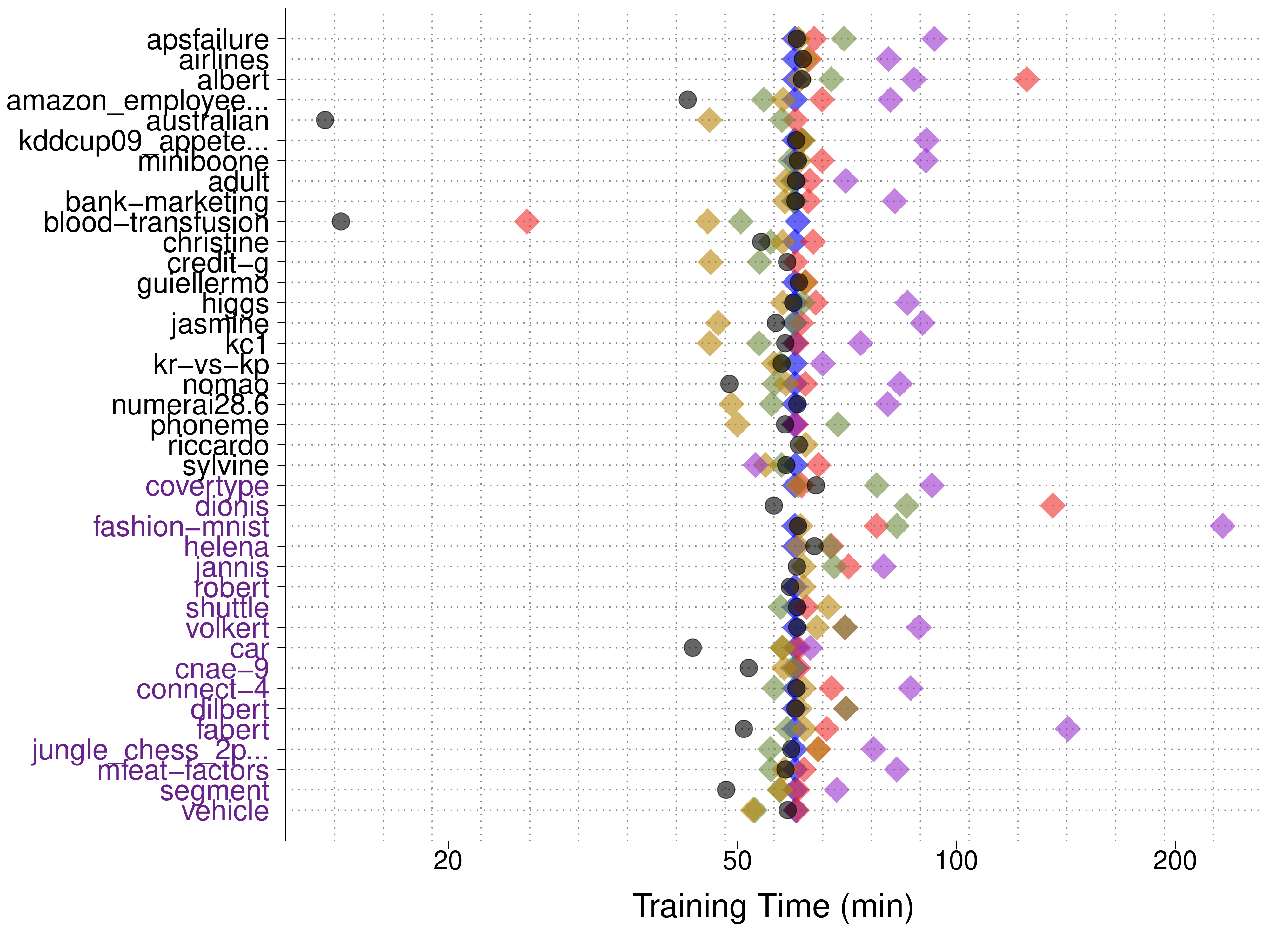}
    &
    \includegraphics[width=0.49\textwidth]{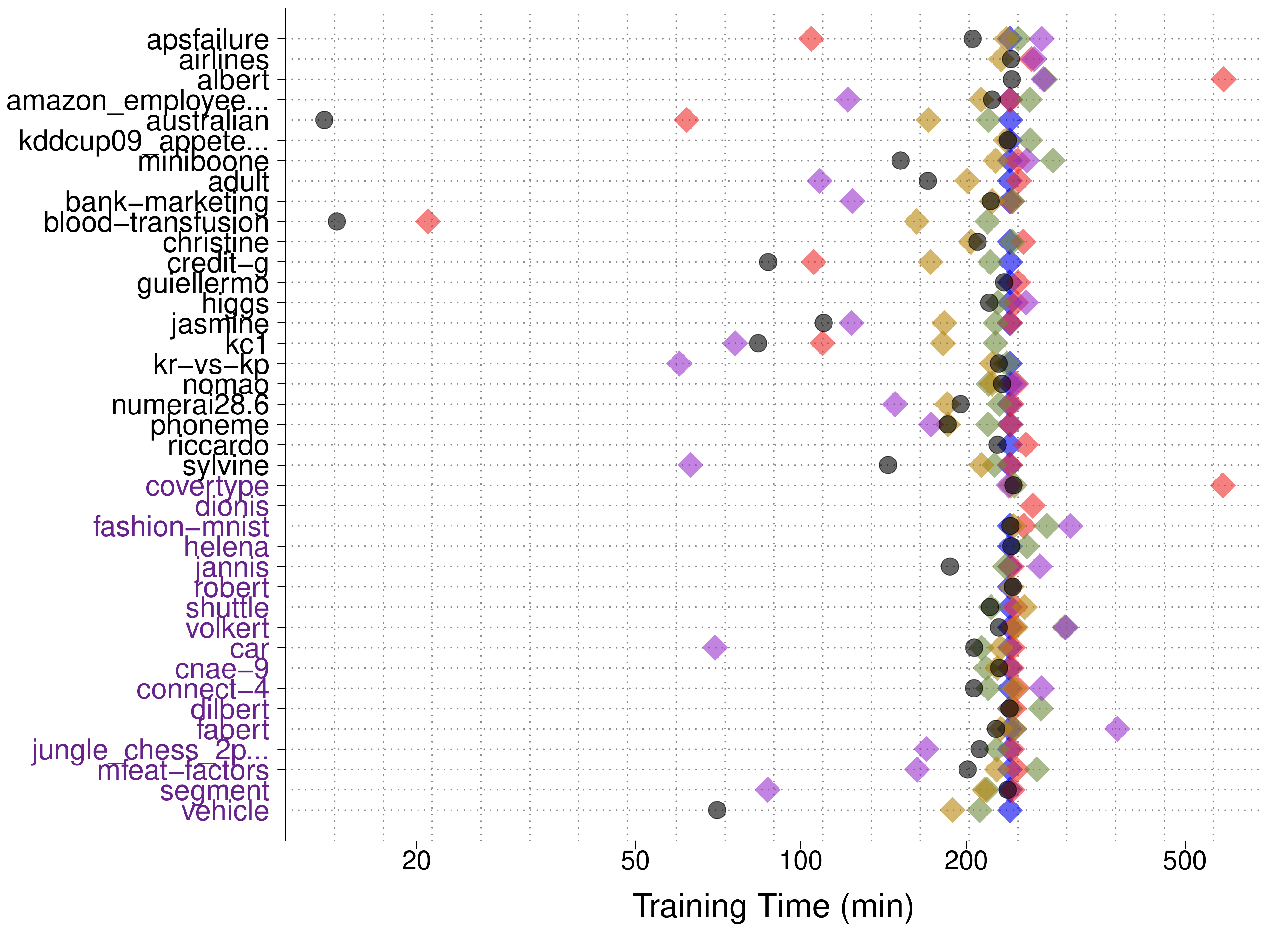} 
    \\ 
    \hspace*{20mm} \textbf{(A)} 1h time limit specified (60 min) 
    &
\hspace*{20mm} \textbf{(B)} 4h time limit specified (240 min) 
\end{tabular}
    \caption{Actual training times of each framework in the AutoML Benchmark, which varied despite the fact that we instructed each framework to only run  for the listed time limit. Unlike AutoGluon, some frameworks vastly exceeded their training time allowance (TPOT in particular).  
    In these cases, the accuracy values presented in this paper presumably  represent optimistic estimates of the performance that would be achieved if training were actually halted at the time limit. Datasets are colored based on  whether they correspond to a binary (black) or multi-class (purple) classification problem.
    }
    \label{fig:openmltime}
\end{figure*}

\begin{table}[tb] 
\centering
\caption{
Ablation analysis of AutoGluon trained without various components on the AutoML Benchmark under 1h and 4h time limits. The ablated variants of AutoGluon are defined in \S\ref{sec:ablation}, columns are defined as in Table \ref{tab:pair_openml_4h} (averaged columns are relative to this table and should not be compared across tables).  Even after 4h, the \emph{NoMultiStack} variant cannot outperform the full AutoGluon trained for only 1h.
}
\label{tab:ablation_combined_4h}
\vskip 0.1in
\begin{tabular}{lcccccc}
\toprule
\textbf{Framework} &  \textbf{Wins} &  \textbf{Losses} &  \textbf{Champion} &  \textbf{Avg. Rank} &  \textbf{Avg. Rescaled Loss} & \textbf{Avg. Time (min)} \\
\midrule
    AutoGluon (4h) &              0 &                0 &                 15 &              2.6757 &                       0.1416 &                      192 \\
     NoRepeat (4h) &             17 &               20 &                 11 &              3.3919 &                       0.2106 &                      114 \\
    AutoGluon (1h) &              5 &               30 &                  2 &              4.6351 &                       0.3323 &                       55 \\
 NoMultiStack (4h) &              7 &               28 &                  3 &              4.6622 &                       0.4361 &                      173 \\
     NoRepeat (1h) &              5 &               32 &                  2 &              4.9595 &                       0.3600 &                       43 \\
 NoMultiStack (1h) &              6 &               31 &                  2 &              6.1351 &                       0.5868 &                       53 \\
        NoBag (1h) &              5 &               33 &                  1 &              6.6351 &                       0.6513 &                       15 \\
        NoBag (4h) &              5 &               33 &                  1 &              7.0405 &                       0.6605 &                       27 \\
    NoNetwork (1h) &              4 &               34 &                  0 &              7.4189 &                       0.7475 &                       10 \\
    NoNetwork (4h) &              5 &               33 &                  0 &              7.4459 &                       0.7431 &                       16 \\
\bottomrule
\end{tabular}

\vskip 0.1in
\end{table}

\iffalse % moved to main text
\begin{table*}[bt]
\centering
\caption{Comparing accuracy of AutoML frameworks after 1h training vs.\ 4h training on the 39 AutoML Benchmark datasets. We count how many times the 1h variant performs better ($>$), worse ($<$), or comparably ($=$) to the 4h variant. AutoGluon and TPOT most reliably demonstrate performance increases going from 1h to 4h. 
}
% \label{tab:pair_openml_1h_vs_4h}
\vspace*{1em}
 \begin{footnotesize}
% \input{openml_core_1h_vs_4h_pairwise_1h_vs_4h.tex}
 \end{footnotesize}
% \end{center}
\vskip -0.1in
\end{table*}
\fi

\begin{table*}[bt]
\centering
\caption{
Comparing the open-source AutoML frameworks over all 10 train/test splits of the AutoML Benchmark (with 4h training time limit). These results are based on the average performance across all 10 folds. A framework failure on any of the 10 folds is considered an overall failure for the dataset. Columns are defined as in Table \ref{tab:pair_openml_4h}, where  averaged columns are relative to the particular table and should not be compared across tables. AutoGluon outperforms \emph{all} other frameworks in 27 of the 38 datasets  (\texttt{Dionis} dataset is excluded from this table because all frameworks failed on this massive dataset). When comparing with all 10 folds, we note AutoGluon has an even better rank and rescaled loss than when evaluating on only the first fold (Table \ref{tab:pair_openml_4h}). Evaluating over 10 folds reduces variance and thus frameworks are less likely to get a strong/poor result by chance.
}
\label{tab:pair_openml_core10fold_4h}
\vspace*{1em}
 \begin{footnotesize}
\begin{tabular}{lccccccc}
\toprule
\textbf{Framework} &  \textbf{Wins} &  \textbf{Losses} &  \textbf{Failures} &  \textbf{Champion} &  \textbf{Avg. Rank} &  \textbf{Avg. Rescaled Loss} & \textbf{Avg. Time (min)} \\
\midrule
         AutoGluon &              0 &                0 &                  1 &                 27 &              1.3684 &                       0.0303 &                      197 \\
        H2O AutoML &              7 &               23 &                  9 &                  6 &              2.4737 &                       0.0955 &                      224 \\
      auto-sklearn &              4 &               28 &                  7 &                  3 &              2.9474 &                       0.1589 &                      240 \\
              TPOT &              3 &               27 &                  9 &                  2 &              3.3158 &                       0.2093 &                      236 \\
         Auto-WEKA &              1 &               31 &                  7 &                  0 &              4.8947 &                       0.9902 &                      242 \\
\bottomrule
\end{tabular}

 \end{footnotesize}
% \end{center}
\vskip -0.1in
\end{table*}

\begin{table*}[bt]
\centering
\caption{
Comparing open-source AutoML frameworks on all 10 folds of the AutoML Benchmark (with 4h training time limit). We include scores reported from the original AutoML Benchmark, indicated with (O). These results are based on the average performance across all 10 folds. A framework failure on any of the 10 folds is considered an overall failure for the dataset. Columns are defined as in Table \ref{tab:pair_openml_4h}, where the averaged columns are relative to the particular table and should not be compared across tables. AutoGluon outperforms \emph{all} other frameworks in 24 of the 39 datasets. Even without access to the original feature type information which was provided in the original benchmark, AutoGluon is still able to outperform the other frameworks. Our runs of the other AutoML frameworks perform similarly to their original results, indicating feature type information can be inferred effectively in most cases.  Note that the original runs failed fewer times than our runs. This is likely because  
% the frameworks are not as stable when given inferred feature types rather than correct feature types, and 
the original AutoML Benchmark runs performed multiple retries of failed frameworks in an attempt to get a result, which we did not consider here.
}
\label{tab:pair_openml_orig10fold_4h}
\vspace*{1em}
 \begin{footnotesize}
\begin{tabular}{lccccccc}
\toprule
\textbf{Framework} &  \textbf{Wins} &  \textbf{Losses} &  \textbf{Failures} &  \textbf{Champion} &  \textbf{Avg. Rank} &  \textbf{Avg. Rescaled Loss} & \textbf{Avg. Time (min)} \\
\midrule
         AutoGluon &              0 &                0 &                  1 &                 24 &              1.8889 &                       0.0391 &                      195 \\
    H2O AutoML (O) &              8 &               29 &                  2 &                  4 &              3.4444 &                       0.0972 &                      208 \\
        H2O AutoML &              7 &               23 &                  9 &                  2 &              3.5000 &                       0.0851 &                      223 \\
  auto-sklearn (O) &              6 &               31 &                  1 &                  2 &              4.6667 &                       0.1385 &                      246 \\
      auto-sklearn &              4 &               28 &                  7 &                  2 &              4.7778 &                       0.1427 &                      240 \\
          TPOT (O) &              7 &               29 &                  3 &                  5 &              4.7778 &                       0.1519 &                      247 \\
              TPOT &              3 &               27 &                  9 &                  0 &              5.3889 &                       0.1949 &                      237 \\
     Auto-WEKA (O) &              1 &               33 &                  4 &                  0 &              8.2222 &                       0.8284 &                      237 \\
         Auto-WEKA &              1 &               31 &                  7 &                  0 &              8.3333 &                       0.7194 &                      242 \\
\bottomrule
\end{tabular}

 \end{footnotesize}
% \end{center}
\vskip -0.1in
\end{table*}

\begin{table*}[bt]
\centering
\caption{
Comparison of the AutoML frameworks on the AutoML Benchmark, evaluating the accuracy metric (with 1h training time limit). Columns are defined as in Table \ref{tab:pair_openml_4h}. Rescaled misclassification rate is calculated in the same manner as rescaled loss, but applied specifically to the accuracy metric. Here, we additionally compare with the commercial Sagemaker Autopilot framework described in \S\ref{sec:autopilot}. All frameworks were optimized on AUC except for AutoPilot, which was optimized on accuracy. This demonstrates that while AutoGluon performs the best in the primary evaluation metric, it also performs favourably on secondary metrics such as accuracy, even compared to AutoPilot which optimized directly on accuracy.
}
\label{tab:pair_openml_accuracy_1h}
\vspace*{1em}
 \begin{footnotesize}
\begin{tabular}{lccccccc}
\toprule
\textbf{Framework} &  \textbf{Wins} &  \textbf{Losses} &  \textbf{Failures} &  \textbf{Champion} &  \textbf{Avg. Rank} &  \textbf{Avg. Rescaled Misclassification Rate} & \textbf{Avg. Time (min)} \\
\midrule
         AutoGluon &              0 &                0 &                  0 &                 17 &              1.8421 &                       0.0509 &                       56 \\
        GCP-Tables &              6 &               15 &                 13 &                  5 &              2.9211 &                       0.1973 &                       83 \\
      auto-sklearn &              7 &               22 &                  5 &                  3 &              3.7105 &                       0.2506 &                       60 \\
        H2O AutoML &              5 &               27 &                  1 &                  1 &              4.0263 &                       0.3198 &                       58 \\
              TPOT &              5 &               26 &                  4 &                  3 &              4.6053 &                       0.4102 &                       67 \\
         AutoPilot &              2 &               23 &                 12 &                  0 &              5.1842 &                       0.4937 &                       58 \\
         Auto-WEKA &              6 &               27 &                  4 &                  4 &              5.7105 &                       0.7212 &                       60 \\
\bottomrule
\end{tabular}

 \end{footnotesize}
% \end{center}
\vskip -0.1in
\end{table*}

\FloatBarrier
\subsection{Additional Results for Kaggle Benchmark}

% LEADERBOARD DIFFERENCE
\begin{figure*}[tb] \centering
\hspace*{5mm} \includegraphics[width=0.55\textwidth]{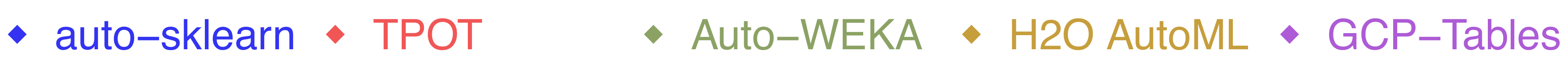} \\
\begin{tabular}{cc}
    \includegraphics[width=0.49\textwidth]{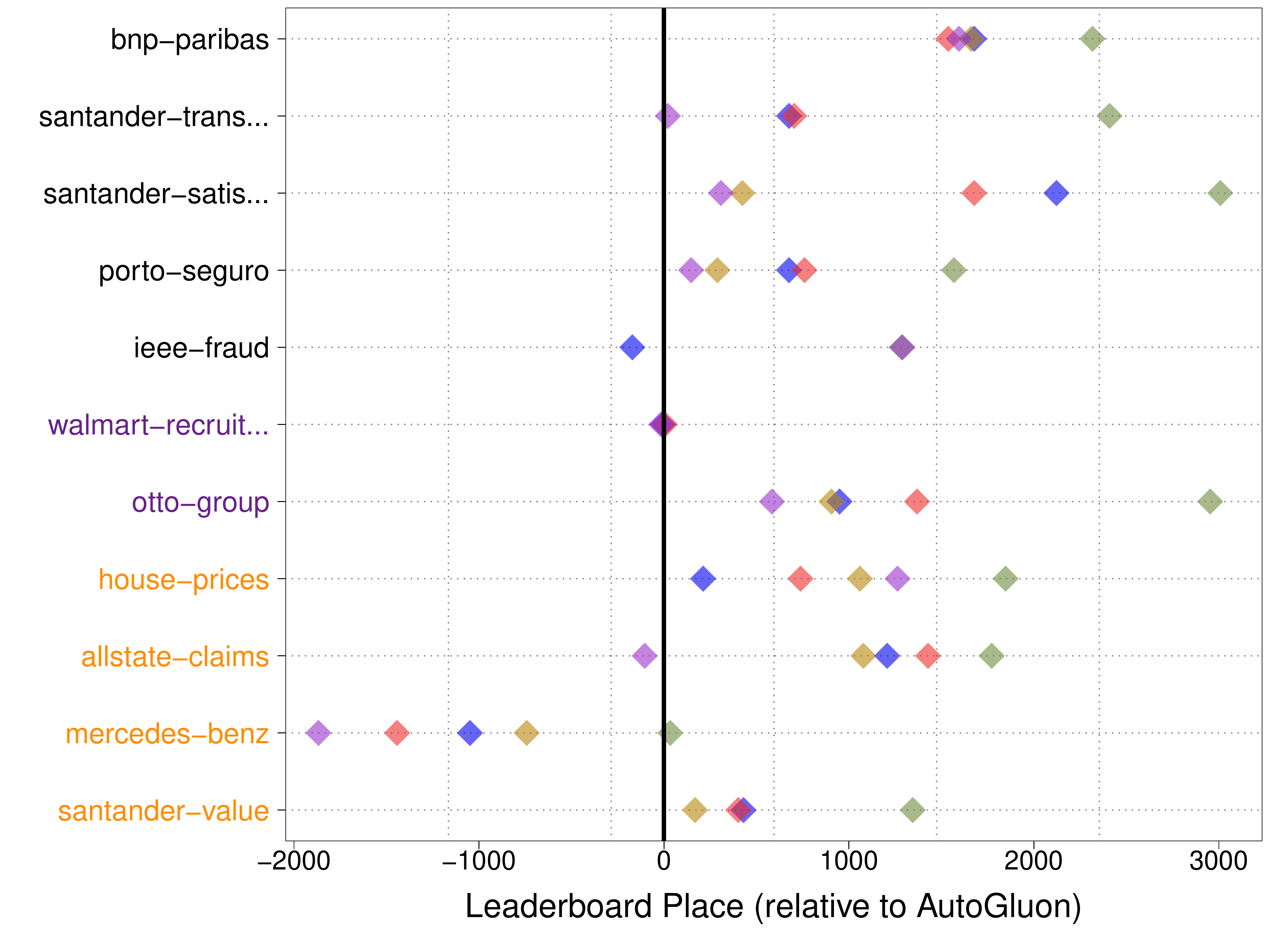}
    &
    \includegraphics[width=0.49\textwidth]{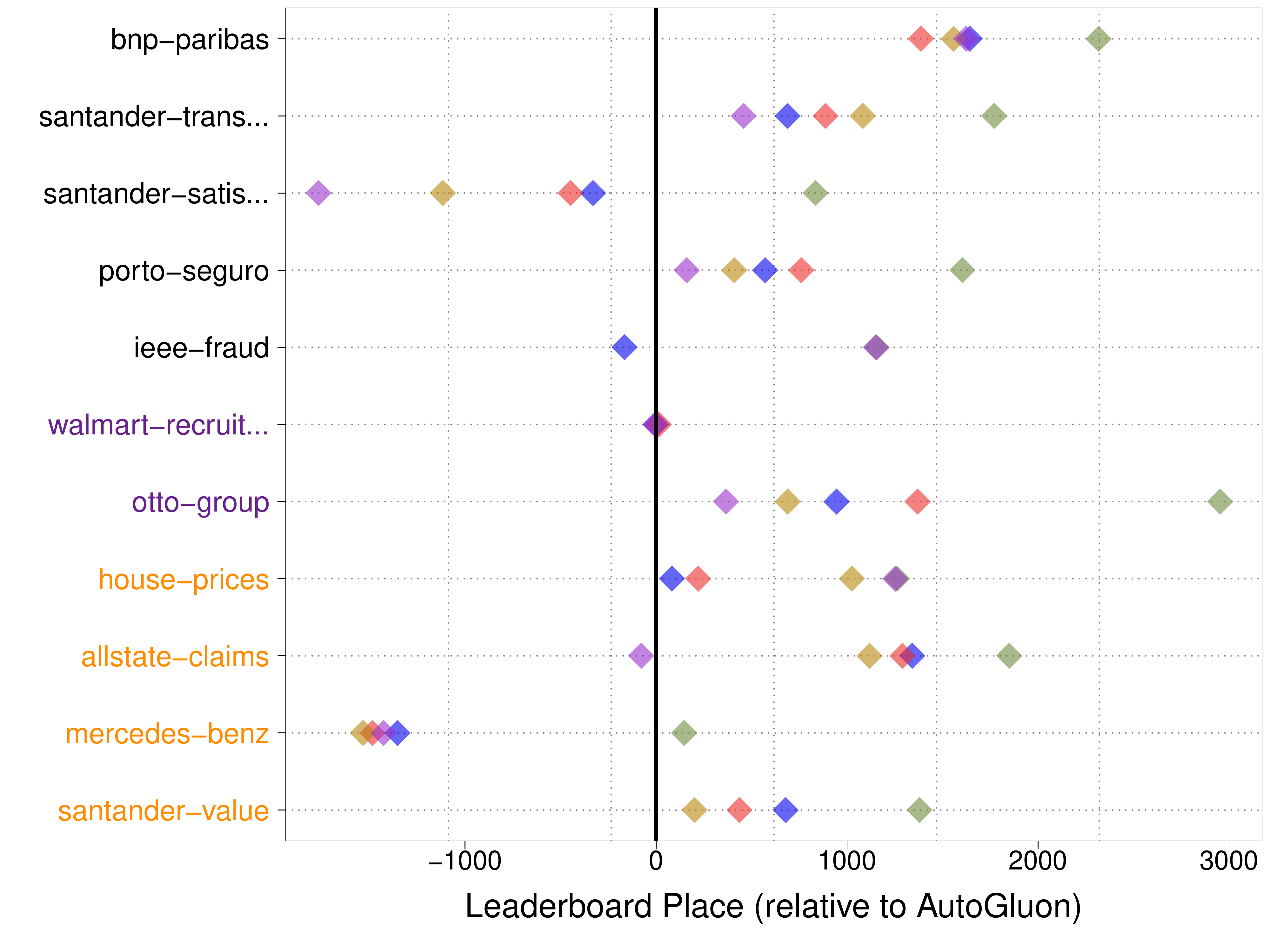}
    \\ 
    \hspace*{20mm} \textbf{(A)} 4h time limit 
    &
\hspace*{20mm} \textbf{(B)} 8h time limit 
\\[1em]
\includegraphics[width=0.49\textwidth]{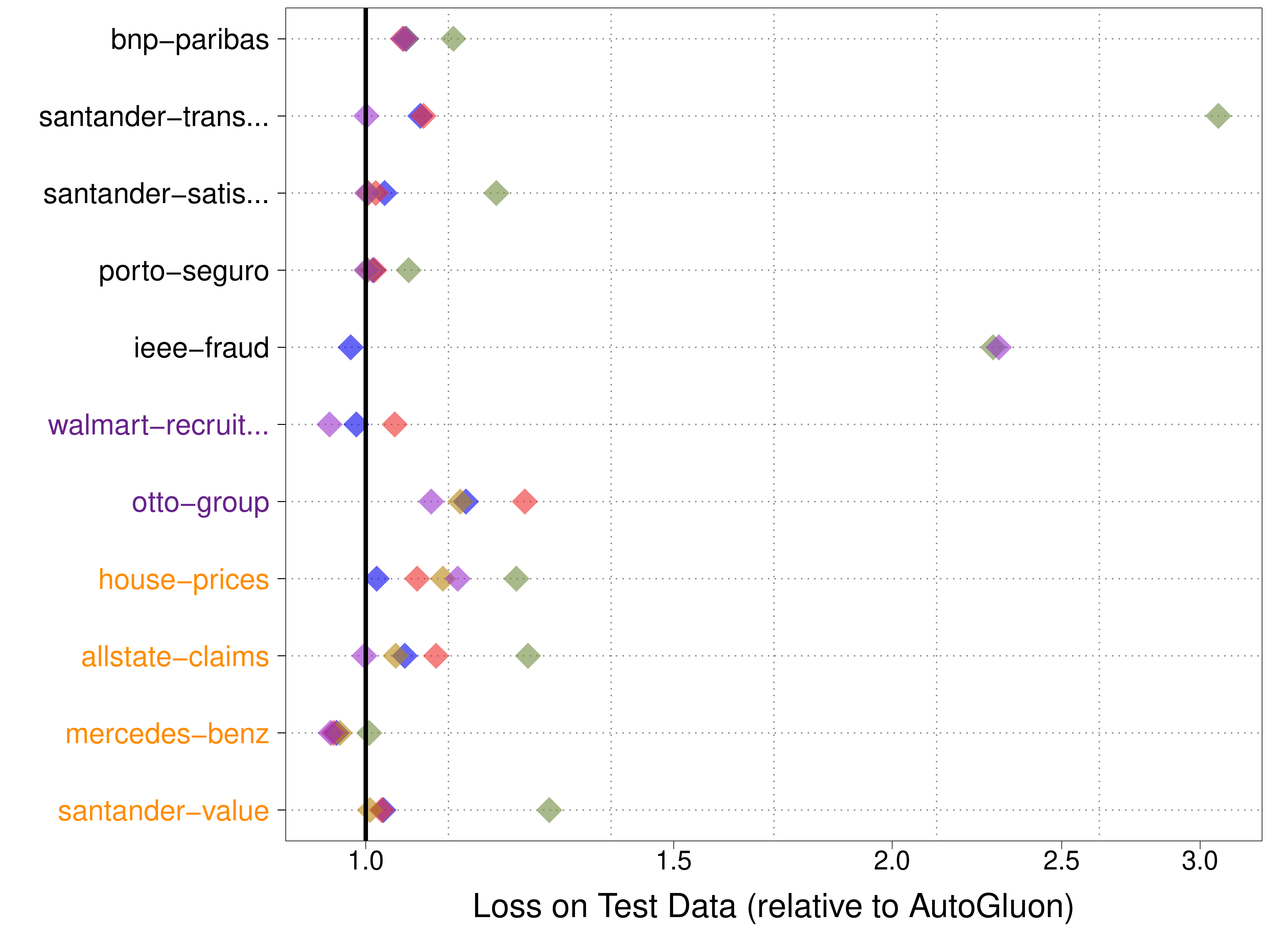}
    &
    \includegraphics[width=0.49\textwidth]{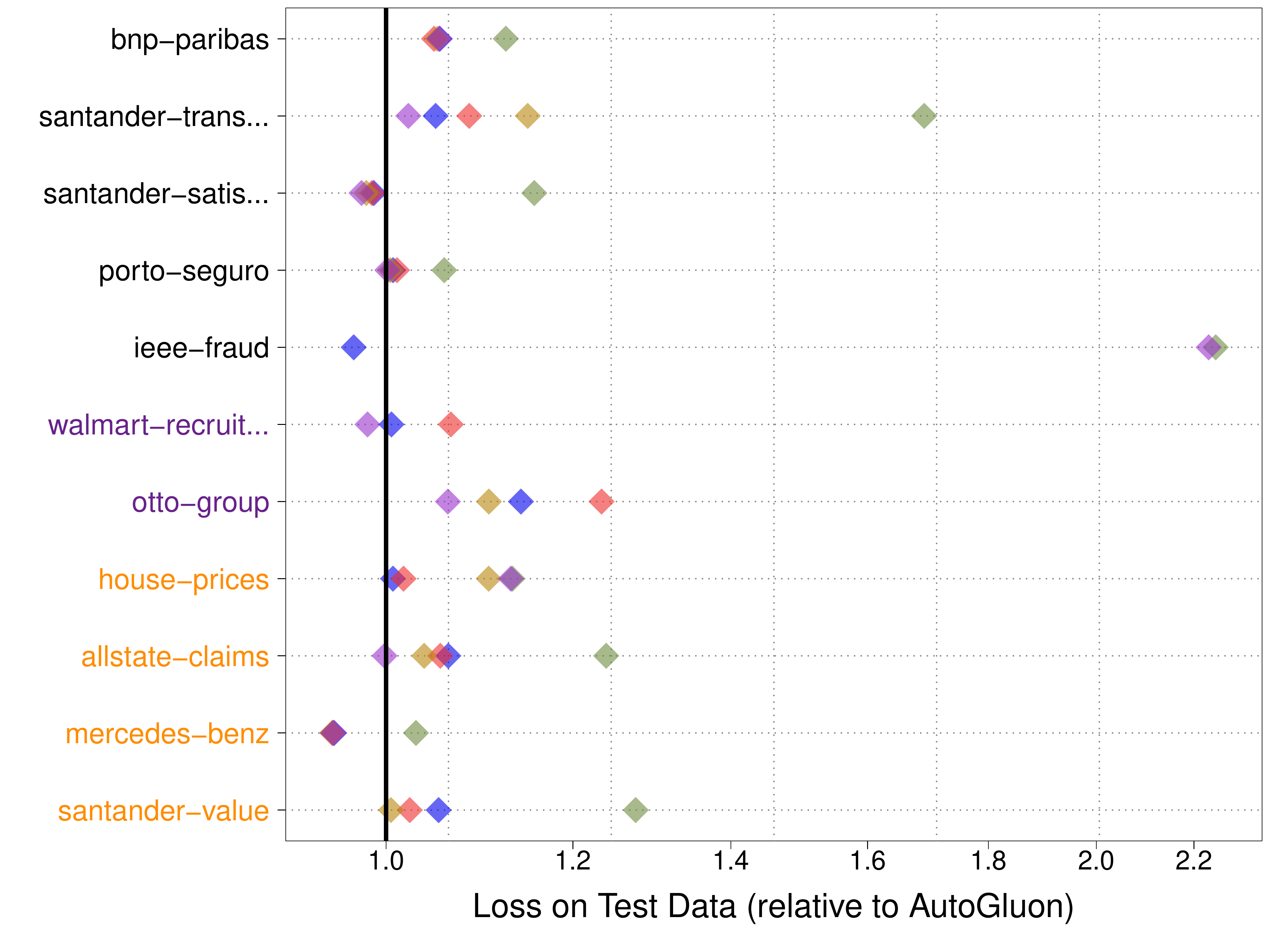} 
    \\ 
    \hspace*{20mm} \textbf{(C)} 4h time limit 
    &
\hspace*{20mm} \textbf{(D)} 8h time limit 
\end{tabular}
 \caption{\textbf{(A)}-\textbf{(B)}: Difference in leaderboard ranks achieved by each AutoML framework vs.\  AutoGluon in the Kaggle competitions (under listed training time limit). These values quantify how much better one framework is vs.\  another, in terms of how many data scientists could beat one but not the other. 
   \textbf{(C)}-\textbf{(D)}: Ratio of loss achieved by AutoML frameworks vs.\ AutoGluon loss on each Kaggle competition (under listed training time limit). Loss is a competition-specific metric (e.g.\ RMSLE, $1-\text{Gini}$, etc.).
   In both plots, points $> 0$ indicate worse performance than AutoGluon and failed runs are not shown. The color of each dataset name indicates the task: binary classification (black), multi-class classification (purple), regression (orange).
    }
    \label{fig:kaggleperf}
\end{figure*}

% TRAINING TIMES KAGGLE
\begin{figure*}[tb] \centering
\hspace*{5mm} \includegraphics[width=0.58\textwidth]{LegendPlot.pdf} \\
\begin{tabular}{cc}
    \includegraphics[width=0.49\textwidth]{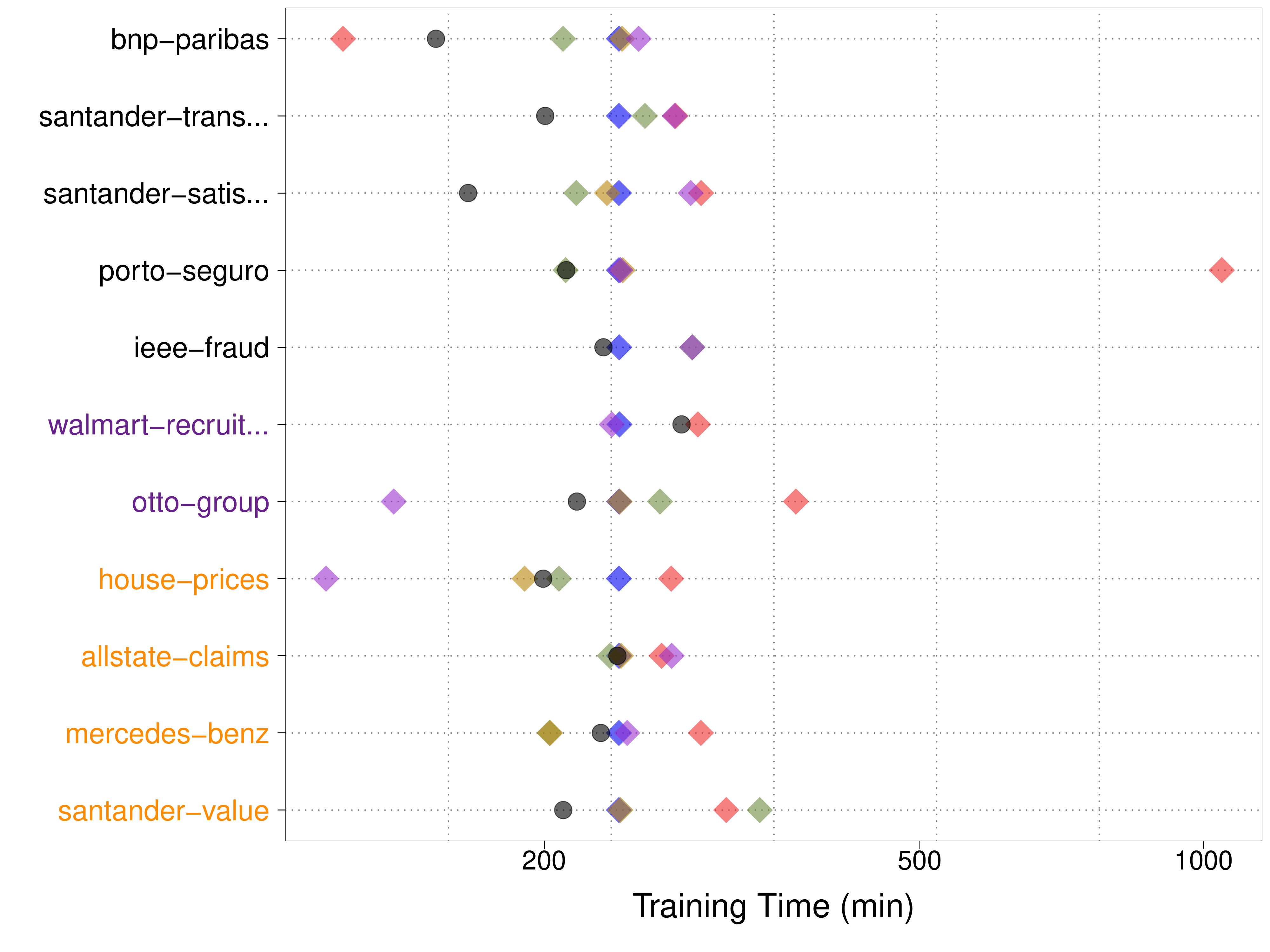}
    &
    \includegraphics[width=0.49\textwidth]{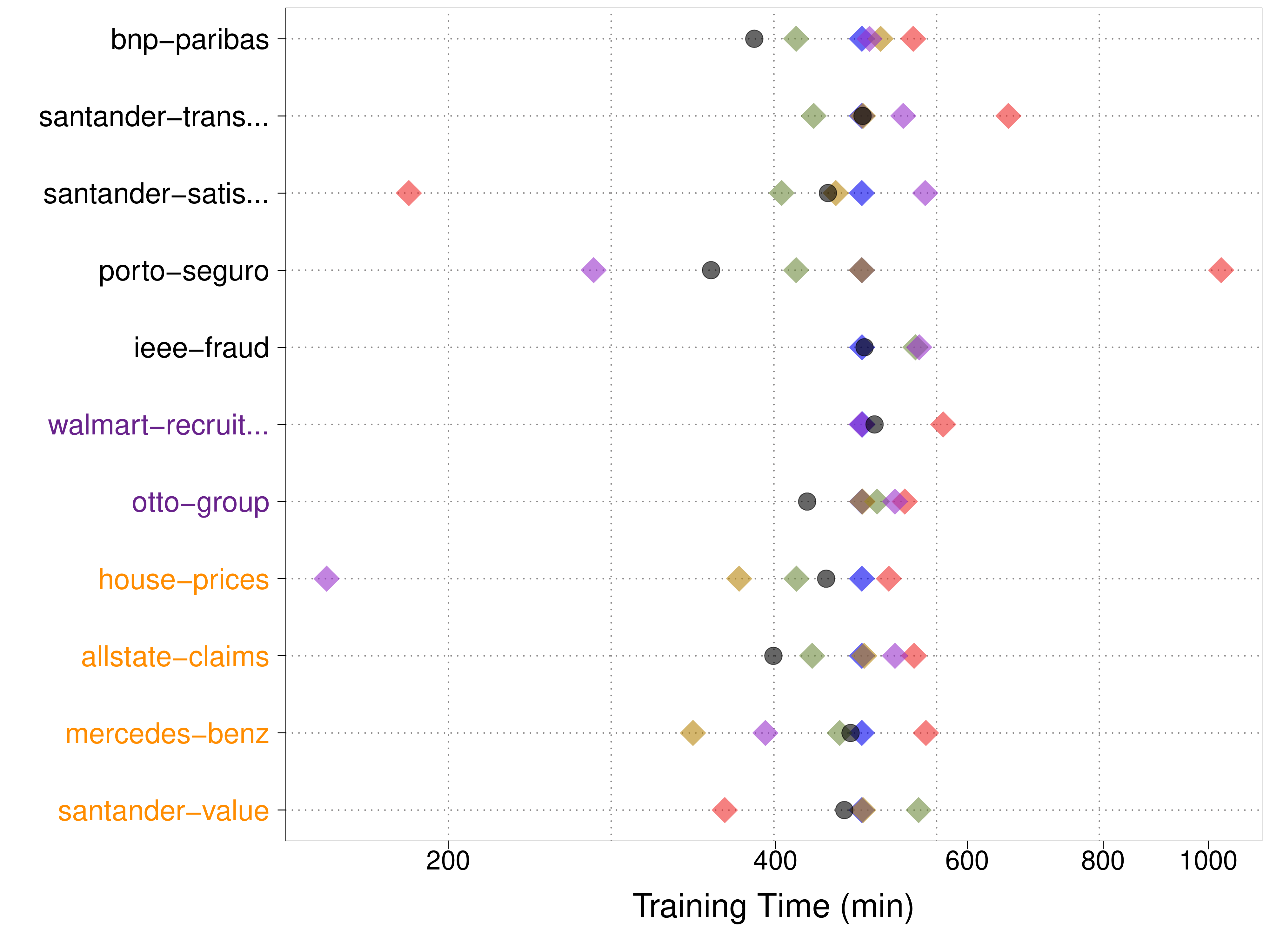} 
    \\ 
    \hspace*{20mm} \textbf{(A)} 4h time limit specified (240 min) 
    &
\hspace*{20mm} \textbf{(B)} 8h time limit specified (480 min) 
\end{tabular}
    \caption{Actual training times of each framework in the Kaggle Benchmark, which varied despite the fact that we instructed each framework to only run for the listed time limit.  Unlike AutoGluon, some frameworks vastly exceeded their training time allowance (TPOT in particular).  In these cases, the accuracy values presented in this paper presumably  represent optimistic estimates of the performance that would be achieved if training were actually halted at the time limit.  The color of each dataset name indicates the corresponding task: binary classification (black), multi-class classification (purple), regression (orange).
    }
    \label{fig:kaggletime}
\end{figure*}
% While some tools only loosely consider this specified time limit, AutoGluon never significantly exceeds it.

\begin{table*}[bth!]
\centering
\caption{Comparing each AutoML framework against AutoGluon on the 11 Kaggle competitions (under 8h time limit). 
Listed are the number of datasets where each framework produced: better predictions than AutoGluon (Wins), worse predictions (Losses), a system failure during training (Failures), or more accurate predictions than all of the other 5 frameworks (Champion). The latter 3 columns show the average: rank of the framework (among the 6 AutoML tools applied to each dataset), percentile rank achieved in on the competition leaderboard (higher = better), and actual training time.  Averages are computed over only the subset of 8 competitions where all methods ran successfully. 
}
\label{tab:pairkag8h}
\vspace*{1em}

 \begin{footnotesize}
\begin{tabular}{lccccccc}
\toprule
\textbf{Framework} &  \textbf{Wins} &  \textbf{Losses} &  \textbf{Failures} &  \textbf{Champion} &  \textbf{Avg. Rank} &  \textbf{Avg. Percentile} & \textbf{Avg. Time (min)} \\
\midrule
         AutoGluon &              0 &                0 &                  0 &                  6 &              2.1250 &                    0.6176 &                      425 \\
        GCP-Tables &              4 &                6 &                  1 &                  3 &              2.5000 &                    0.5861 &                      426 \\
        H2O AutoML &              2 &                7 &                  2 &                  1 &              3.0000 &                    0.5068 &                      448 \\
              TPOT &              2 &                8 &                  1 &                  0 &              3.5000 &                    0.4793 &                      565 \\
      auto-sklearn &              3 &                8 &                  0 &                  1 &              3.8750 &                    0.4851 &                      480 \\
         Auto-WEKA &              0 &               10 &                  1 &                  0 &              6.0000 &                    0.2161 &                      435 \\
\bottomrule
\end{tabular}

 \end{footnotesize}
% \end{center}
\vskip -0.1in
\end{table*}

\begin{figure*}[h!] \centering
\includegraphics[width=0.52\textwidth]{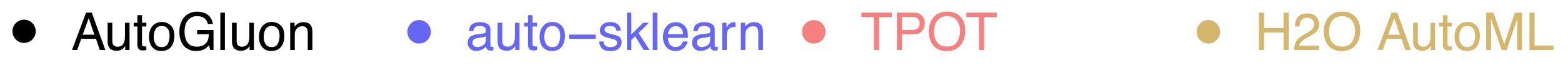} 
\\
    \includegraphics[width=0.57\textwidth]{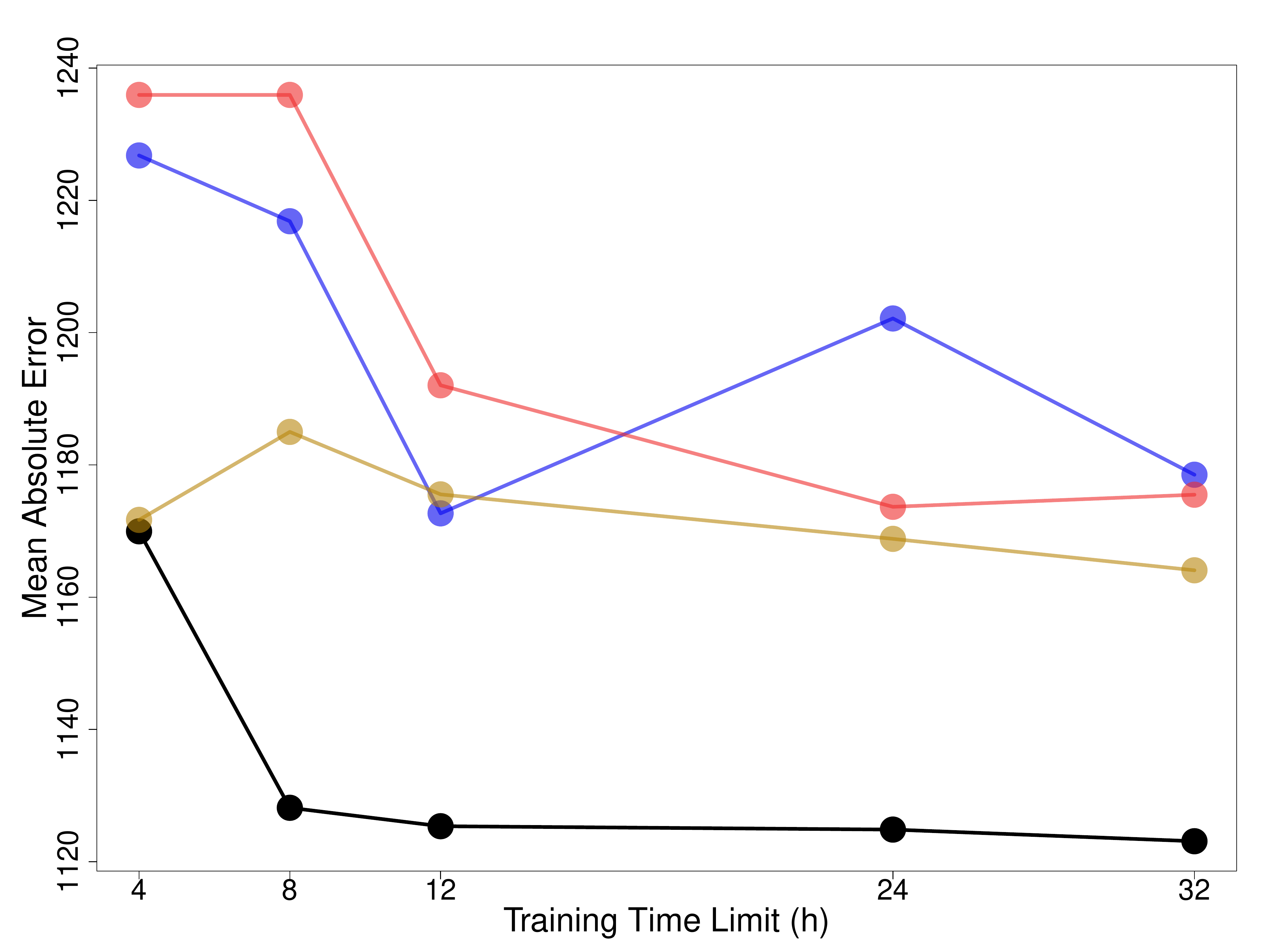}
    \caption{Predictive performance of open-source AutoML frameworks under different training time limits (Auto-WEKA not shown as it exhibited outlying poor performance). 
    Here we show the example of the \texttt{allstate-claims}  dataset from the Kaggle Benchmark, which is a regression task evaluated via the mean absolute error metric. Unlike the other frameworks, AutoGluon performance consistently improves with longer time limits.
    }
    \label{fig:trainingtimesallstate}
\end{figure*}

\begin{table}[bt]
\centering
\caption{Performance of AutoML frameworks after 4h training vs.\ 8h training on the 11 datasets of the Kaggle Benchmark. We count how many times the 4h variant performs better ($>$), worse ($<$), or comparably ($=$) to the 8h variant. 
}
\label{tab:kaggle_4h_vs_8h}
\vspace*{1em}
 \begin{footnotesize}
\textit{}\begin{tabular}{lccc}
\toprule
\textbf{System} &  \textbf{$>$ 8h} &  \textbf{$<$ 8h} &  \textbf{$=$ 8h} \\
\midrule
      AutoGluon &                4 &                7 &                0 \\
     GCP-Tables &                6 &                4 &                0 \\
     H2O AutoML &                2 &                5 &                1 \\
   auto-sklearn &                3 &                8 &                0 \\
           TPOT &                2 &                5 &                3 \\
      Auto-WEKA &                3 &                2 &                5 \\
\bottomrule
\end{tabular}

 \end{footnotesize}
% \end{center}
\vskip -0.1in
\end{table}

\FloatBarrier

\subsection{Complete Set of Performance Numbers}

This section lists the performance of each AutoML framework on every dataset from our benchmarks.

% RAW OPENML RESULTS

\begin{table*}[!bth] % 4h all Kaggle percentiles
\centering
\caption{Loss on test data produced by AutoML frameworks in our AutoML Benchmark (after training with 1h time limit).
The best performance among all AutoML frameworks is highlighted in bold, and failed runs are indicated by a cross.
}
\label{tab:rawlossopenml1h}
\vspace*{1em}

 \begin{footnotesize}
\begin{tabular}{lcccccc}
\toprule
  \textbf{Dataset} & \textbf{Auto-WEKA} & \textbf{auto-sklearn} &   \textbf{TPOT} & \textbf{H2O AutoML} & \textbf{GCP-Tables} & \textbf{AutoGluon} \\
\midrule
        APSFailure &              0.047 &                 0.006 &           0.011 &               0.008 &               0.005 &     \textbf{0.004} \\
          Airlines &                 x  &                 0.276 &           0.309 &      \textbf{0.267} &               0.271 &              0.291 \\
            Albert &              0.334 &                 0.250 &           0.294 &      \textbf{0.238} &               0.242 &              0.238 \\
 Amazon\_employee\_a &              0.180 &                 0.129 &  \textbf{0.118} &               0.120 &               0.129 &              0.118 \\
        Australian &     \textbf{0.045} &                    x  &           0.076 &               0.060 &                  x  &              0.050 \\
         Covertype &              1.070 &                 0.113 &           0.605 &               0.326 &               0.197 &     \textbf{0.058} \\
            Dionis &              1.277 &                    x  &           2.545 &                  x  &                  x  &     \textbf{0.964} \\
     Fashion-MNIST &              1.254 &                 0.389 &           0.792 &               0.300 &      \textbf{0.273} &              0.286 \\
            Helena &              9.134 &                 2.887 &           3.039 &               2.697 &                  x  &     \textbf{2.615} \\
            Jannis &              2.302 &                    x  &           0.736 &               0.676 &               0.676 &     \textbf{0.661} \\
 KDDCup09\_appetenc &              0.271 &                 0.167 &           0.179 &               0.182 &               0.154 &     \textbf{0.145} \\
         MiniBooNE &              0.084 &                 0.015 &           0.019 &               0.015 &               0.014 &     \textbf{0.014} \\
            Robert &                 x  &                 1.714 &              x  &      \textbf{1.563} &                  x  &              1.606 \\
           Shuttle &       \textbf{0.0} &                 0.000 &           0.000 &               0.001 &                  x  &              0.000 \\
           Volkert &              2.114 &                 0.919 &           1.005 &               0.837 &               0.839 &     \textbf{0.727} \\
             adult &              0.093 &                 0.069 &           0.071 &               0.069 &               0.068 &     \textbf{0.068} \\
    bank-marketing &              0.079 &                 0.061 &           0.063 &               0.061 &               0.061 &     \textbf{0.059} \\
 blood-transfusion &              0.236 &        \textbf{0.225} &           0.268 &               0.241 &                  x  &              0.231 \\
               car &              0.721 &                 0.005 &           0.002 &      \textbf{4e-05} &               0.000 &              0.000 \\
         christine &              0.231 &                 0.161 &           0.178 &               0.160 &                  x  &     \textbf{0.155} \\
            cnae-9 &              0.393 &                 0.087 &  \textbf{0.057} &               0.112 &                  x  &              0.145 \\
         connect-4 &              0.812 &                 0.470 &           0.358 &               0.350 &      \textbf{0.309} &              0.328 \\
          credit-g &              0.145 &                    x  &  \textbf{0.135} &               0.159 &                  x  &              0.154 \\
           dilbert &              0.515 &                 0.044 &           0.226 &               0.070 &                  x  &     \textbf{0.027} \\
            fabert &              2.514 &                 0.750 &           0.884 &               0.726 &               0.785 &     \textbf{0.652} \\
        guiellermo &                 x  &                 0.114 &           0.134 &               0.091 &                  x  &     \textbf{0.074} \\
             higgs &              0.355 &                 0.189 &           0.196 &               0.183 &      \textbf{0.179} &              0.182 \\
           jasmine &              0.133 &        \textbf{0.118} &           0.142 &               0.125 &               0.134 &              0.130 \\
 jungle\_chess\_2pcs &              0.618 &                 0.203 &           0.201 &               0.238 &      \textbf{0.007} &              0.025 \\
               kc1 &              0.141 &                 0.180 &           0.158 &               0.159 &               0.166 &     \textbf{0.135} \\
          kr-vs-kp &              0.000 &          \textbf{0.0} &              x  &               0.000 &               8e-05 &              4e-05 \\
     mfeat-factors &              0.122 &                 0.127 &           0.122 &               0.104 &               0.104 &     \textbf{0.074} \\
             nomao &              0.007 &        \textbf{0.001} &           0.002 &               0.003 &               0.004 &              0.002 \\
       numerai28.6 &              0.476 &                 0.476 &              x  &               0.477 &      \textbf{0.475} &              0.484 \\
           phoneme &              0.050 &                 0.037 &           0.036 &               0.034 &               0.033 &     \textbf{0.024} \\
          riccardo &                 x  &                    x  &              x  &      \textbf{0.000} &                  x  &     \textbf{0.000} \\
           segment &              0.239 &                 0.075 &           0.066 &               0.068 &               0.063 &     \textbf{0.047} \\
           sylvine &              0.021 &                 0.013 &  \textbf{0.008} &               0.015 &               0.020 &              0.013 \\
           vehicle &              0.817 &                 0.363 &           0.315 &      \textbf{0.303} &                  x  &              0.311 \\
\bottomrule
\end{tabular}

 \end{footnotesize}
% \end{center}
\vskip -0.1in
\end{table*}

\begin{table*}[!bth] % 4h all Kaggle percentiles
\centering
\caption{Loss on test data produced by AutoML frameworks in our AutoML Benchmark (after training with 4h time limit). 
The best performance among all AutoML frameworks is highlighted in bold, and failed runs are indicated by a cross.
}
\label{tab:rawlossopenml4h}
\vspace*{1em}

 \begin{footnotesize}
\begin{tabular}{lcccccc}
\toprule
  \textbf{Dataset} & \textbf{Auto-WEKA} & \textbf{auto-sklearn} &   \textbf{TPOT} & \textbf{H2O AutoML} & \textbf{GCP-Tables} & \textbf{AutoGluon} \\
\midrule
        APSFailure &              0.031 &                 0.008 &           0.011 &               0.008 &               0.012 &     \textbf{0.006} \\
          Airlines &                 x  &                    x  &           0.309 &      \textbf{0.267} &               0.270 &              0.287 \\
            Albert &              0.333 &                    x  &           0.294 &                  x  &               0.241 &     \textbf{0.231} \\
 Amazon\_employee\_a &              0.125 &                 0.139 &  \textbf{0.112} &               0.117 &               0.125 &              0.113 \\
        Australian &              0.048 &        \textbf{0.042} &           0.067 &               0.060 &                  x  &              0.050 \\
         Covertype &              1.070 &                    x  &           0.547 &               0.242 &               0.086 &     \textbf{0.056} \\
            Dionis &                 x  &                    x  &  \textbf{2.546} &                  x  &                  x  &                 x  \\
     Fashion-MNIST &              1.254 &                 0.343 &           0.778 &               0.256 &               0.263 &     \textbf{0.238} \\
            Helena &              6.335 &                 2.723 &              x  &                  x  &                  x  &     \textbf{2.454} \\
            Jannis &              1.692 &                 0.685 &           0.726 &                  x  &               0.663 &     \textbf{0.658} \\
 KDDCup09\_appetenc &              0.206 &                 0.164 &              x  &               0.163 &                  x  &     \textbf{0.149} \\
         MiniBooNE &              0.083 &                 0.015 &           0.018 &               0.015 &               0.014 &     \textbf{0.012} \\
            Robert &                 x  &                 1.435 &              x  &               1.338 &                  x  &     \textbf{1.278} \\
           Shuttle &       \textbf{0.0} &                 0.000 &           0.000 &               0.001 &                  x  &              0.000 \\
           Volkert &              6.968 &                 0.853 &           0.930 &               0.811 &               0.772 &     \textbf{0.706} \\
             adult &                 x  &                 0.068 &           0.069 &               0.070 &               0.068 &     \textbf{0.067} \\
    bank-marketing &              0.076 &                 0.069 &           0.063 &               0.061 &               0.062 &     \textbf{0.059} \\
 blood-transfusion &              0.236 &                    x  &           0.287 &               0.269 &                  x  &     \textbf{0.231} \\
               car &       \textbf{0.0} &                 0.000 &           0.005 &               9e-05 &               0.000 &       \textbf{0.0} \\
         christine &              0.201 &                 0.154 &           0.175 &               0.159 &                  x  &     \textbf{0.151} \\
            cnae-9 &              0.393 &                 0.303 &           0.092 &      \textbf{0.088} &                  x  &              0.135 \\
         connect-4 &              0.811 &                 0.437 &           0.360 &               0.334 &      \textbf{0.297} &              0.326 \\
          credit-g &              0.309 &                 0.181 &           0.184 &               0.147 &                  x  &     \textbf{0.137} \\
           dilbert &              0.515 &                 0.062 &           0.158 &               0.043 &                  x  &     \textbf{0.021} \\
            fabert &              1.129 &                 0.751 &           0.826 &               0.710 &               0.772 &     \textbf{0.649} \\
        guiellermo &                 x  &                 0.091 &           0.090 &                  x  &                  x  &     \textbf{0.070} \\
             higgs &              0.212 &                 0.185 &           0.195 &                  x  &      \textbf{0.177} &              0.180 \\
           jasmine &              0.125 &                 0.114 &  \textbf{0.113} &               0.125 &               0.139 &              0.130 \\
 jungle\_chess\_2pcs &              0.933 &                 0.192 &           0.103 &                  x  &      \textbf{0.003} &              0.011 \\
               kc1 &              0.160 &                    x  &           0.148 &               0.156 &               0.168 &     \textbf{0.134} \\
          kr-vs-kp &              0.000 &          \textbf{0.0} &              x  &               4e-05 &               0.000 &              4e-05 \\
     mfeat-factors &              0.124 &                 0.091 &  \textbf{0.059} &               0.106 &               0.121 &              0.066 \\
             nomao &              0.007 &        \textbf{0.001} &           0.001 &               0.003 &               0.003 &     \textbf{0.001} \\
       numerai28.6 &              0.476 &                 0.477 &           0.478 &               0.477 &      \textbf{0.472} &              0.486 \\
           phoneme &              0.035 &                 0.039 &           0.025 &               0.034 &               0.032 &     \textbf{0.023} \\
          riccardo &                 x  &        \textbf{0.000} &           0.000 &                  x  &                  x  &              0.000 \\
           segment &              0.239 &                 0.068 &           0.060 &               0.068 &               0.071 &     \textbf{0.047} \\
           sylvine &              0.018 &                 0.016 &  \textbf{0.006} &               0.014 &               0.021 &              0.011 \\
           vehicle &              1.278 &                 0.403 &              x  &               0.310 &                  x  &     \textbf{0.308} \\
\bottomrule
\end{tabular}

 \end{footnotesize}
% \end{center}
\vskip -0.1in
\end{table*}

\iffalse % TODO: just to ensure forward references align in submission
% Accuracy
\begin{table*}[!bth] % 4h all Kaggle percentiles
\centering
\caption{Misclassification Error Rate ($= 1 -$ Accuracy) on test data produced by AutoML frameworks in the AutoML Benchmark (after training with 1h time limit). 
The best performance among all AutoML frameworks is highlighted in bold, and failed runs are indicated by a cross. This table contains AutoPilot as an additional AutoML framework tested (see \S\ref{sec:autopilot}). Note that all frameworks were optimized for the AUC objective (as in the original AutoML Benchmark), except for AutoPilot which was optimized directly for accuracy.
}
\label{tab:rawlossopenmlaccuracy1h}
\vspace*{1em}

 \begin{footnotesize}
\input{openml_accuracy_1h_openml_alllosses_1h.tex}
 \end{footnotesize}
% \end{center}
\vskip -0.1in
\end{table*}
\fi

% RAW KAGGLE RESULTS
\begin{table*}[!bth] % 4h all Kaggle percentiles
\centering
\caption{Percentile ranks on each competition leaderboard achieved by various AutoML frameworks in our Kaggle Benchmark (training with 4h time limit).  
The best performance among all AutoML frameworks is highlighted in bold, and failed runs are indicated by a cross.
}
\label{tab:percentileskagg4h}
\vspace*{1em}

 \begin{footnotesize}
\begin{tabular}{lcccccc}
\toprule
\textbf{Dataset} & \textbf{Auto-WEKA} & \textbf{auto-sklearn} & \textbf{TPOT} & \textbf{H2O AutoML} & \textbf{GCP-Tables} & \textbf{AutoGluon} \\
\midrule
           house &              0.420 &                 0.748 &         0.643 &               0.578 &               0.537 &     \textbf{0.791} \\
        mercedes &              0.160 &                 0.444 &         0.547 &               0.363 &      \textbf{0.658} &              0.169 \\
           value &              0.114 &                 0.319 &         0.325 &               0.377 &                  x  &     \textbf{0.415} \\
        allstate &              0.124 &                 0.310 &         0.237 &               0.352 &      \textbf{0.740} &              0.706 \\
     bnp-paribas &              0.193 &                 0.412 &         0.460 &               0.417 &               0.440 &     \textbf{0.986} \\
     transaction &              0.131 &                 0.329 &         0.326 &                  x  &               0.404 &     \textbf{0.406} \\
    satisfaction &              0.235 &                 0.408 &         0.495 &               0.740 &               0.763 &     \textbf{0.823} \\
           porto &              0.158 &                 0.331 &         0.315 &               0.406 &               0.434 &     \textbf{0.462} \\
      ieee-fraud &              0.119 &        \textbf{0.349} &            x  &                  x  &               0.119 &              0.322 \\
         walmart &                 x  &                 0.390 &         0.379 &                  x  &      \textbf{0.398} &              0.384 \\
            otto &              0.145 &                 0.717 &         0.597 &               0.729 &               0.821 &     \textbf{0.988} \\
\bottomrule
\end{tabular}

 \end{footnotesize}
% \end{center}
\vskip -0.1in
\end{table*}

\begin{table*}[!bth] % 8h all Kaggle percentiles
\centering
\caption{Percentile ranks on each competition leaderboard achieved by various AutoML frameworks in our Kaggle Benchmark (training with 8h time limit). 
The best performance among all AutoML frameworks is highlighted in bold, and failed runs are indicated by a cross.
}
\label{tab:percentileskagg8h}
\vspace*{1em}

 \begin{footnotesize}
\begin{tabular}{lcccccc}
\toprule
\textbf{Dataset} & \textbf{Auto-WEKA} & \textbf{auto-sklearn} & \textbf{TPOT} & \textbf{H2O AutoML} & \textbf{GCP-Tables} & \textbf{AutoGluon} \\
\midrule
           house &              0.531 &                 0.767 &         0.740 &               0.578 &               0.533 &     \textbf{0.784} \\
        mercedes &              0.114 &                 0.507 &         0.541 &      \textbf{0.554} &               0.526 &              0.153 \\
           value &              0.114 &                 0.271 &         0.325 &               0.377 &                  x  &     \textbf{0.423} \\
        allstate &              0.124 &                 0.291 &         0.308 &               0.364 &      \textbf{0.757} &              0.731 \\
     bnp-paribas &              0.193 &                 0.423 &         0.511 &               0.453 &               0.430 &     \textbf{0.986} \\
     transaction &              0.225 &                 0.349 &         0.326 &               0.304 &               0.375 &     \textbf{0.427} \\
    satisfaction &              0.235 &                 0.463 &         0.486 &               0.617 &      \textbf{0.744} &              0.399 \\
           porto &              0.158 &                 0.359 &         0.322 &               0.390 &               0.438 &     \textbf{0.469} \\
      ieee-fraud &              0.118 &        \textbf{0.326} &            x  &                  x  &               0.119 &              0.300 \\
         walmart &                 x  &                 0.392 &         0.379 &                  x  &      \textbf{0.397} &              0.393 \\
            otto &              0.145 &                 0.718 &         0.597 &               0.791 &               0.883 &     \textbf{0.988} \\
\bottomrule
\end{tabular}

 \end{footnotesize}
% \end{center}
\vskip -0.1in
\end{table*}

\FloatBarrier
\section{AutoML Failures}
\label{sec:errors}

When designing AutoML solutions, it is not easy to ensure stability and robustness across all manner of datasets that may be encountered in the wild. During our benchmarking process, we encountered various errors and defects in the tested frameworks. These errors include out-of-memory errors, resource limit errors, column name formatting errors, frameworks failing to generate models, frameworks failing to finish in their allotted time limit (and never stopping), data type inference errors, errors due to too few training rows, errors due to too many features, errors due to too many classes, output formatting errors, test data rows randomly shuffled during predictions, and many more.  The results presented in this paper represent our best effort to resolve and mitigate as many of these errors as possible. 

Many existing AutoML frameworks attempt to simultaneously train many models in parallel on different CPUs, but this leads to memory issues when working with large datasets. In contrast, AutoGluon simply trains each model one at a time, preferring to leverage all available CPUs to reduce individual model-training times as much as possible. This makes a big difference in practice.  Both auto-sklearn and Auto-WEKA train models in parallel, and exhibited numerous memory issues when we ran them on larger datasets with less powerful CPUs, despite the fact that AutoGluon worked fine in these settings.

\subsection{Failures in AutoML Benchmark}

Below is a breakdown detailing what types of errors each AutoML framework encountered in the AutoML Benchmark.  Oddly, certain frameworks would error on particular train/test splits while succeeding on other train/test splits of the same dataset.

\subsubsection{AutoGluon Failures}

AutoGluon failed on 1 of the 39 datasets in the 4h runs (0 failures in the 1h runs).  This dataset is Dionis. Dionis is a large dataset of 416,188 rows, 61 features, and 355 classes.  We note that no framework succeeded on all 10 folds of Dionis except for AutoGluon with 1h time limit. Both of the commercial AutoML services we tried, GCP-Tables and AutoPilot, also failed to produce a result for Dionis.

Due to AutoGluon's multi-layer stacking, it generates 355 features per successful base model to use as input to the stackers. While most of the models succeed or properly catch memory errors before they happen, the AutoGluon Neural Network does not yet have such a safeguard in the version of AutoGluon used in our benchmarks.  Therefore, AutoGluon would randomly succeed or fail a fold depending on how much memory the neural network ended up using. AutoGluon 4h succeeded on folds 0 and 1, but failed on fold 2 with an out-of-memory error that prevented the sequential completion of further folds. Because not all 10 folds were ran, we report a total failure for AutoGluon on this dataset. 

\subsubsection{GCP-Tables Failures}

GCP-Tables failed on 14 of the 39 datasets.

\begin{enumerate}

\item  \texttt{ValueError: GCP AutoML tables can only be trained on datasets with >= 1000 rows}

GCP-Tables has a limitation of requiring at least 1000 rows of training data. This caused failures on 5 datasets: Australian, blood-transfusion, cnae-9, credit-g, and vehicle.

\item  \texttt{GoogleAPICallError: None Too many columns: XXXX. Maximum number is: 1000}

GCP-Tables has a limitation of requiring no more than 1000 features. This caused failures on 5 datasets: christine, dilbert, guiellermo, riccardo, Robert.

\item  \texttt{AssertionError: GCP AutoML did not predict with all classes! GCP returned 40 of XXX classes!}

GCP-Tables appears to only return 40 classes' prediction probabilities on multi-class classification problems with greater than 40 classes, despite being directly given log-loss as the evaluation metric to optimize for. Because not all class probabilities were returned, the log-loss would have been infinite, and thus we consider this a failure. This caused failures on 2 datasets: Helena, Dionis.

\item  \texttt{GoogleAPICallError: None Missing label(s) in test split: target column contains 7 distinct values, but only 6 present. There must be at least one instance of each label value in every split.}

GCP-Tables failed on 1 dataset with this error: Shuttle. We suspect this is due to Shuttle having its least frequent class appear only 9 times in the training set, and GCP-Tables attempted to use 10\% of the training data as test data which contained 0 instances of this rare class, causing the crash.

\item  \texttt{GoogleAPICallError: None INTERNAL}

GCP-Tables cryptically failed on 1 dataset, KDDCup09\_appetency, despite training for the full 4h duration.

\end{enumerate}

\subsubsection{H2O AutoML Failures}

H2O AutoML failed on 9 of the 39 datasets. Note that the errors listed here only account for the 4 hour runs.

\begin{enumerate}

\item  \texttt{H2OConnectionError: Local server has died unexpectedly. RIP.}

This error occurred on several of the larger datasets, and often only on a fraction of folds. It is a cryptic error and likely represents a large variety of potential root causes. This error occurred on 7 datasets: Albert, guiellermo, higgs, Jannis, jungle\_chess\_2pcs\_raw\_endgame\_complete, KDDCup09\_appetency, and riccardo.

\item  \texttt{AssertionError: H2O could not produce any model in the requested time.}

This error occurred on 1 dataset: Dionis.

\item  \texttt{H2O trains far longer than requested}

This error occurred on 1 dataset: Helena. On 5 of the 10 folds, H2O trained for approximately 90,000 seconds (25 hours), compared to the requested 4 hours. It is unknown why H2O only appears to have acted this way on one dataset and only on half of the folds, nor why it stopped training rather sharply at 90,000 seconds.

\end{enumerate}

\subsubsection{auto-sklearn Failures}

auto-sklearn failed on 7 of the 39 datasets. Note that the errors listed here only account for the 4 hour runs.

\begin{enumerate}

\item  \texttt{auto-sklearn hard crashes with SegmentationFault}

This error occurred on 5 datasets: Airlines, Albert, blood-transfusion, Covertype, and kc1. While Airlines, Albert, and Covertype are all very large datasets where out-of-memory is a likely error reason, blood-transfusion is the smallest dataset in the benchmark, and is therefore an odd dataset to fail on for this reason. Furthermore, all 5 of these datasets did not hard crash in the 1h time limit runs. auto-sklearn had 0 such failures in the 1 hour runs. This indicates that auto-sklearn ran out of memory or ran out of disk space (750 GB disk allocated per run) by training too many or too large of models. This would explain the failure on blood-transfusion, given that auto-sklearn trains between 7,000 and 10,000 models in a single hour on blood-transfusion.

\item  \texttt{ValueError: attempt to get argmin of an empty sequence}

This error indicates that auto-sklearn did not finish training any models. This error occurred on 1 dataset: Dionis.

\item  \texttt{AssertionError: found prediction probability value outside of [0, 1]!}

This error indicates that auto-sklearn somehow created a model which outputted a probability value outside of valid bounds. This error occurred on 1 dataset: phoneme. This interestingly only occurred on a single fold of phoneme, with all others succeeding.

\end{enumerate}

\subsubsection{TPOT Failures}

TPOT failed on 9 of the 39 datasets. Note that the errors listed here only account for the 4 hour runs.

\begin{enumerate}

\item  \texttt{TPOT never finishes training}

TPOT does not always respect time limits, and in some cases appears to take a far greater time to train or may even get permanently stuck. For these results, we gave each algorithm up to 3 times the allocated time to finish, and these datasets were still running for TPOT. Several of these runs continued to train for weeks without an indication of stopping. This error occurred on 4 datasets: Helena, KDDCup09\_appetency, kr-vs-kp, and vehicle.

\item  \texttt{RuntimeError: A pipeline has not yet been optimized. Please call fit() first.}

This error occurs when TPOT has not finished training any models in the allocated time. This error occurred on 2 datasets: Dionis and Robert. An interesting note is that while Robert trained for well over the requested time (Averaging 21000 seconds), Dionis failed with this error on average in only 500 seconds, indicating that some internal error occurred with no models being fit.

\item  \texttt{RuntimeError: The fitted pipeline does not have the predict\_proba() function.}

This is a cryptic error due to TPOT being explicitly passed the AUC and log-loss evaluation metrics for binary and multi-class classification respectively. It appears that occasionally TPOT will construct an invalid pipeline which it selects as its final solution. This is likely a defect internally in TPOT, and only happens on a fraction of folds and seemingly at random. This error occurred on 3 datasets: credit-g, numerai28.6, and riccardo.

\end{enumerate}

\subsubsection{Auto-WEKA Failures}

Auto-WEKA failed on 7 of the 39 datasets. Note that the errors listed here only account for the 4 hour runs.

\begin{enumerate}

\item  \texttt{Auto-WEKA hard crashes with SegmentationFault}

Auto-WEKA does not safely handle memory in all instances, and this causes a hard-crash that prohibits the return of the exact exception message. This error occurred on 6 datasets: adult, Airlines, Dionis, guiellermo, riccardo, and Robert.

\item  \texttt{MemoryError: Unable to allocate 2.70 GiB for an array with shape (522912, 99) and data type <U14}

This memory error occurred on 1 dataset: Covertype.

\item  \texttt{ValueError: AutoWEKA failed producing any prediction.}

This memory error occurred on 1 dataset: Covertype. Note that Covertype had different errors depending on the fold, with 5 of the 10 folds succeeding.

\end{enumerate}

\subsubsection{AutoPilot Failures}

AutoPilot failed on 12 of the 39 datasets.

\begin{enumerate}

\item  \texttt{AssertionError: Could not complete the data builder processing job. The AutoML Job cannot continue. Failed Job Arn: arn:aws:sagemaker:XXX...}

Upon further inspection into the log files of these failed jobs, it is revealed that, like GCP-Tables, AutoPilot requires a minimum of 1000 rows of training data, and the datasets that failed in this manner all have less than 1000 training rows. This error occurred on 5 datasets: Australian, blood-transfusion, cnae-9, credit-g, and vehicle.

\item  \texttt{AssertionError: AutoPilot did not finish training any models}

AutoPilot failed to finish training any models in the allocated time. This error occurred on 5 datasets: Covertype, Fashion-MNIST, guiellermo, riccardo, and Robert.

\item  \texttt{KeyError: "None of [Index(['false', 'true'], dtype='object')] are in the [columns]"}

AutoPilot inferred labels with string values 'true' and 'false' to be 1 and 0 respectively. Upon returning the predictions, they were in the form 1 and 0, despite all other string type label values returning their original string names in the other datasets. Because the values were given as 'true' and 'false', but returned as '1' and '0', automatically processing these results will cause a crash to many systems attempting to use AutoPilot, and thus we consider it an error. This error occurred on 1 dataset: kc1.

\item  \texttt{AssertionError: Could not complete the candidate generation processing job. The AutoML Job cannot continue. Failed Job Arn: arn:aws:sagemaker:XXX...}

This cryptic error was thrown less than 15 minutes into the run and likely indicates that the dataset was too large for the data processing functionality to handle without encountering errors. This error occurred on 1 dataset: Dionis.

\end{enumerate}

\subsection{Failures in Kaggle Benchmark}

Below, we compile the list of AutoML failures observed in the Kaggle Benchmark, which prevented a framework from producing predictions for the corresponding competition. AutoGluon exhibited no errors on any dataset under any of the training time limits specified in this paper.  

\begin{enumerate}

\item H2O failed on the \texttt{ieee-fraud-detection} data with error:

\texttt{java.lang.IllegalArgumentException: Test/Validation dataset has a non-categorical column 'dist1' which is categorical in the training data}

However, these data appear correctly formatted, and all other AutoML frameworks ran successfully in this competition.   A similar H2O error has been  discussed in the Kaggle forums for this competition: \\
\url{https://www.kaggle.com/c/ieee-fraud-detection/discussion/110643}

\item H2O failed on the \texttt{walmart-recruiting-trip-type-classification} data with error:

\texttt{AssertionError: H2O could not produce any model in the requested time.}

Even increasing the allowed training time to 32 hours did not solve this issue.

\item H2O failed on the \texttt{santander-customer-transaction-prediction} data under a 4h time limit, with repeated trials always producing the error:

\texttt{AssertionError: H2O could not produce any model in the requested time.}

We note that 8h time limit was sufficient for H2O to produce predictions for this competition.

\item TPOT failed on the \texttt{ieee-fraud-detection} data with error:

\texttt{RuntimeError: A pipeline has not yet been optimized. Please call fit() first.}

This error message could indicate TPOT has not had enough time to find any valid ML pipelines, but we found even greatly increasing the allowed TPOT runtime limit up to 32h did not solve this issue.

\item Despite being given evaluation metrics that require probabilistic predictions (e.g.\ AUC, Log-Loss) for certain datasets, TPOT nonetheless occasionally failed with error:

\texttt{RuntimeError: The fitted pipeline does not have the predict\_proba() function.}

By re-running TPOT, we managed to circumvent this issue and successfully produce predictions for each of the Kaggle datasets.

\item GCP-Tables could not produce models for the \texttt{santander-value-prediction-challenge} competition because this data contains 4992 columns and GCP-Tables refuses to handle data with over 1000 columns.

\item In some competitions, % (\texttt{ieee-fraud-detection} in particular), 
GCP-Tables occasionally failed to return predictions for every single test data point (presumably producing errors during inference for certain test rows).  Because a prediction must be submitted for every test example in order to get a score from Kaggle, we simply imputed dummy predictions for these missing cases, using: the marginal probability distribution over classes in the training data for classification tasks, and the average $y$-value in the training data for regression tasks.

\item GCP-Tables (8h) failed initially on the \texttt{santander-customer-satisfaction} data with error:

\texttt{google.api\_core.exceptions.GoogleAPICallError: None INTERNAL}

but was able to run successfully when retried.

\item Auto-WEKA failed on the \texttt{walmart-recruiting-trip-type-classification} data with opaque error:

\texttt{java.lang.IllegalArgumentException: A nominal attribute (feature2) cannot have duplicate labels ('(1.384628-1.384628]')}

Note that after the AutoGluon preprocessing (including one-hot encoding of categoricals), all features were declared as numeric in the ARFF files provided to Auto-WEKA. When run for 24h,  Auto-WEKA succeeded on this data, indicating this error is time-limit related.  However, the resulting performance of the 24h Auto-WEKA run was very poor as it predicted certain classes with near-zero probability even though the specified evaluation metric is log-loss. 

\item Auto-WEKA often performed poorly under the log-loss evaluation metric because it occasionally produced predicted probabilities = 0 for certain classes, which are severely penalized under this metric. We added a small $\epsilon = $1e-8 factor to such predictions to ensure finite log-loss values.  Note that Auto-WEKA was always informed the log-loss would be used (via argument:  \texttt{metric = kBInformation}), so why it produced such overconfident predictions is unclear.

\end{enumerate}

\clearpage 
\bibliographystylesi{icml2020}
\bibliographysi{autogluon-automl} % supplement bibliography.

\end{document}